\documentclass[sigconf]{acmart}
\usepackage{url}
\usepackage{amsthm}
\usepackage{afterpage}

\usepackage[normalem]{ulem}
\useunder{\uline}{\ul}{}
\usepackage{graphicx}
\usepackage{textcomp}
\usepackage{xcolor}
\usepackage{balance} 
\usepackage{bbm}
\usepackage{amsmath,amsfonts}

\usepackage{epstopdf}
\usepackage{array}
\usepackage{booktabs}

\usepackage{color}
\usepackage{xcolor}
\usepackage{colortbl}
\usepackage{xspace}
\usepackage{multirow}
\usepackage{pifont}
\usepackage{enumitem}
\usepackage{bbding}
\usepackage{hyperref}
\usepackage{makecell}
\usepackage{microtype}
\usepackage{fontawesome}
\usepackage{caption,setspace}
\usepackage{subfigure}
\usepackage[utf8]{inputenc}
\usepackage{utfsym}
\usepackage{diagbox}
\usepackage{makecell}
\AtBeginDocument{%
 }

\copyrightyear{2025}
\acmYear{2025}
\setcopyright{acmlicensed}
\acmConference[KDD '25]{Proceedings of the 31st ACM SIGKDD Conference on Knowledge Discovery and Data Mining V.2}{August 3--7, 2025}{Toronto, ON, Canada}
\acmBooktitle{Proceedings of the 31st ACM SIGKDD Conference on Knowledge Discovery and Data Mining V.2 (KDD '25), August 3--7, 2025, Toronto, ON, Canada}
\acmDOI{10.1145/3711896.3736956}
\acmISBN{979-8-4007-1454-2/2025/08}

\ccsdesc[500]{Computing methodologies~Distributed artificial intelligence}
\ccsdesc[500]{Computing methodologies~Machine learning}

\keywords{Federated Continual Graph Learning; Graph Neural Networks}

\usepackage[linesnumbered,ruled,vlined]{algorithm2e}

\newcommand{\ssymbol}[1]{^{\@fnsymbol{#1}}}

\SetKwInput{KwInput}{Input}                
\SetKwInput{KwOutput}{Output}              
\SetKwRepeat{Do}{do}{while}
\DeclareMathOperator*{\argmax}{argmax}

\definecolor{darkred}{rgb}{0.55, 0.0, 0.0}
\definecolor{lightred}{rgb}{0.94, 0.5, 0.5}
\definecolor{rosybrown}{rgb}{0.74, 0.56, 0.56}
\definecolor{darkgreen}{rgb}{0.0, 0.39, 0.0}
\definecolor{skyblue}{rgb}{0.56, 0.93, 0.56}
\definecolor{darkblue}{rgb}{0.0, 0.0, 0.55}
\definecolor{lightblue}{rgb}{0.68, 0.85, 0.9}
\definecolor{skyblue}{rgb}{0.53, 0.81, 0.92}

\definecolor{gray}{rgb}{0.5, 0.5, 0.5}
\definecolor{forestgreen}{rgb}{0.13, 0.55, 0.13}
\definecolor{slateblue}{rgb}{0.42, 0.35, 0.80}
\definecolor{darkgray}{rgb}{0.33, 0.33, 0.33}
\definecolor{royalblue}{rgb}{0.25, 0.41, 0.88}

\SetCommentSty{mycommfont}

\usepackage{ulem}

\begin{document}

\title{Federated Continual Graph Learning}

\author{Yinlin Zhu}
\affiliation{%
\institution{Sun Yat-sen University}
\city{Guangzhou}
\country{China}}
\email{zhuylin27@mail2.sysu.edu.cn}

\author{Miao Hu}
\affiliation{%
\institution{Sun Yat-sen University}
\city{Guangzhou}
\country{China}}
\email{humiao5@mail.sysu.edu.cn}

\author{Di Wu}
\authornote{Corresponding author}
\affiliation{
  \institution{Sun Yat-sen University}
  \city{Guangzhou}
  \country{China}
}
\email{wudi27@mail.sysu.edu.cn}

\renewcommand{\shortauthors}{Yinlin Zhu, Miao Hu, Di Wu}

\begin{abstract}
    
Managing evolving graph data presents substantial challenges in storage and privacy, and training graph neural networks (GNNs) on such data often leads to catastrophic forgetting, impairing performance on earlier tasks. Despite existing continual graph learning (CGL) methods mitigating this to some extent, they rely on centralized architectures and ignore the potential of distributed graph databases to leverage collective intelligence. To this end, we propose Federated Continual Graph Learning (FCGL) to adapt GNNs across multiple evolving graphs under storage and privacy constraints. Our empirical study highlights two core challenges: local graph forgetting (LGF), where clients lose prior knowledge when adapting to new tasks, and global expertise conflict (GEC), where the global GNN exhibits sub-optimal performance in both adapting to new tasks and retaining old ones, arising from inconsistent client expertise during server-side parameter aggregation. To address these, we introduce POWER, a framework that preserves experience nodes with maximum local-global coverage locally to mitigate LGF, and leverages pseudo-prototype reconstruction with trajectory-aware knowledge transfer to resolve GEC. Experiments on various graph datasets demonstrate POWER's superiority over federated adaptations of CGL baselines and vision-centric federated continual learning approaches.
    
\end{abstract}

\maketitle
\newcommand\kddavailabilityurl{https://doi.org/10.5281/zenodo.15487654}

\ifdefempty{\kddavailabilityurl}{}{
\begingroup\small\noindent\raggedright\textbf{KDD Availability Link:}\\
The source code and further technical report of this paper have been made publicly available at \url{\kddavailabilityurl}.
\endgroup
}
\section{Introduction}
\label{sec: introduction}

    Graph neural networks (GNNs) have emerged as a powerful framework for harnessing relationships in graph-structured data, enabling superior performance across diverse graph-based AI applications, including recommendation~\cite{xia2023app_gnn_rec1, yang2023app_gnn_rec2}, biomedical~\cite{bang2023app_gnn_bio1, qu2023app_gnn_bio2}, and finance~\cite{balmaseda2023app_gnn_fina1, hyun2023app_gnn_fina2}. However, in the era of big data, graph structures often evolve as new entities and relationships emerge, while storing or accessing massive historical node profiles (e.g., features and labels) and topologies is impractical considering database capacity and security~\cite{qin2023large_dynamic_1}. As a result, GNNs often experience unacceptable performance degradation on past tasks (i.e., graphs prior to changes) when adapting to current tasks, known as the \textit{catastrophic forgetting} problem, a consequence of a weak stability-plasticity tradeoff~\cite{mccloskey1989cata_1, ratcliff1990cata_2}. To tackle this issue, inspired by the success of continual learning in vision tasks~\cite{madaan2021cl_cv1, mallya2018cl_cv2, wang2021cl_cv3}, various continual graph learning (CGL) studies have been proposed in recent years, all of which have demonstrated satisfactory performance~\cite{zhou2021ergnn, liu2021twp, su2023cgl_ssrm_regularization, choi2024dslr, zhang2024pdgnns}. 

    \subsection*{\normalsize{\textit{\ \ \ \ Overlooked Potential: CGL with Collective Intelligence}}}

    Despite their effectiveness, current CGL methods are based on the assumption of centralized data storage, where a single institution collects and manages the entire evolving graph. However, this assumption does not always hold in real-world applications. In many practical scenarios, graph-structured data is often distributed across multiple decentralized sources, each capturing and responding to local changes in nodes and topology over time independently. 
    
    \begin{figure}[t]
      \includegraphics[width=0.45\textwidth]{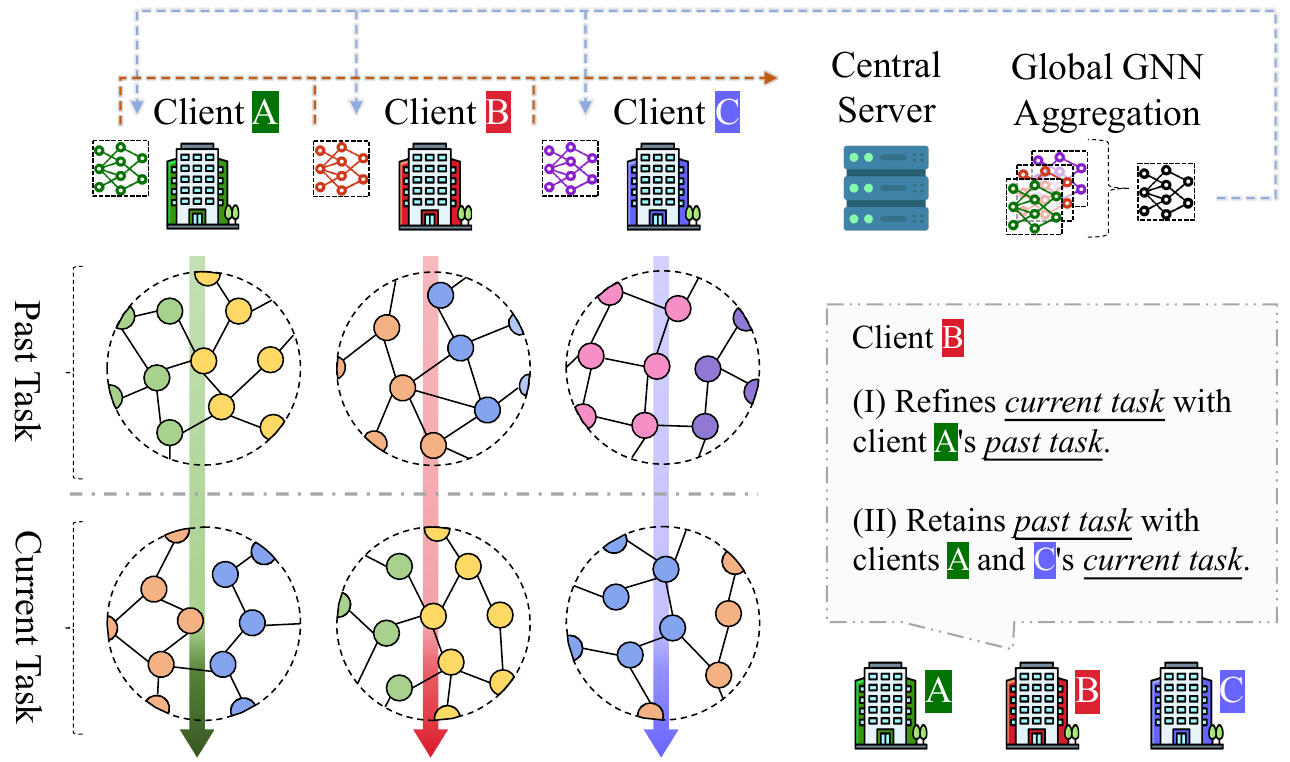}
      \captionsetup{font={small,stretch=1}}
      \caption{\textbf{Illustration of our FCGL paradigm}, which achieves collaborative CGL through multi-round GNN parameter communications between clients and server. Each client's evolving graph is represented by two tasks, with node colors indicating different categories.}
      \label{fig: motivation_fcgl}
      \vspace{-0.45cm}
    \end{figure}

    \textbf{Motivating Scenario.} In the e-commerce domain, various online platforms maintain private evolving product networks, where nodes represent products and edges capture association relationships (e.g., co-purchases or shared reviews), both evolving continuously as new products or associations are introduced~\cite{hu2020ogb, shchur2018amazon_datasets}.
    
    In such decentralized settings, existing CGL methods face \textbf{inherent limitations}, as each institution trains its GNN independently, relying solely on its own evolving graph. This isolated learning differs from human learning, which benefits from collective intelligence through group knowledge sharing (e.g., social media, seminars)~\cite{salminen2012collective}. 
    Obviously, if an institution could leverage knowledge shared by others, it would be better positioned to mitigate catastrophic forgetting and adapt to new tasks. However, constraints like data privacy and commercial competition make directly aggregating multiple evolving graphs unfeasible, presenting significant challenges to achieving effective CGL with collective intelligence.

    \subsection*{\normalsize{\ \ \ \ \textit{First Exploration: Federated Continual Graph Learning}}}

    Building on the success of federated graph learning (FGL)~\cite{zhang2021fedsage}, we introduce federated continual graph learning (FCGL) as the first practical framework for collaborative CGL in decentralized environments. As shown in Fig. \ref{fig: motivation_fcgl}, FCGL employs a multi-round communication process. The clients first train local GNNs on their private evolving graphs, and a central server then aggregates these models to form a global GNN, which has improved performance across diverse local graphs. The global model is subsequently broadcast to the clients, enabling the leveraging of collective intelligence without directly sharing massive and sensitive evolving graphs. Notably, although various federated continual learning methods \cite{dong2022glfc, yoon2021fedweit, zhang2023target, tran2024lander} have demonstrated effectiveness in vision tasks, their algorithms rely on augmentation strategies and network architectures tailored specifically for image data, rendering them inapplicable to graph data. Furthermore, for FCGL, the characteristics of multi-client evolving graphs, feasibility, and potential challenges impacting the GNN performance remain unexplored.

    \subsection*{\normalsize{\ \ \ \ \textit{Our Efforts: Establishing an Effective FCGL Paradigm}}}
    \label{sec: Our Efforts}
    
    To establish an effective FCGL paradigm, we make these efforts:

    \textbf{E1: In-depth Empirical Investigation of FCGL.} 
    We conduct in-depth empirical investigations of FCGL (Sec.~\ref{sec: empirical investigation}). From the 
    \textcolor{darkred}{\textbf{Data}} perspective, we explore the data distribution of multi-client evolving graphs and identify the phenomenon of divergent graph evolution trajectories, which leads each client to develop expertise in specific classes at different stages, forming the foundation of their collective intelligence. From the \textcolor{darkgreen}{\textbf{Feasibility}} perspective, we observe that even the simplest implementation of FCGL outperforms isolated CGL methods in decentralized scenarios, demonstrating its ability to harness collective knowledge from multi-client evolving graphs. From the \textcolor{darkblue}{\textbf{Effectiveness}} perspective, we identify two non-trivial challenges impacting GNN performance, including (1) \underline{\textit{Local Graph Forgetting}} (LGF), where the local GNNs forget the knowledge from the previous tasks when adapting to new ones, and (2) \underline{\textit{Global Expertise Conflict}} (GEC), where the global GNN performs sub-optimally on both past and current tasks due to graph knowledge conflicts, which occur during parameter aggregation across clients with divergent graph evolution trajectories.

    \textbf{E2: Novel and Effective FCGL Framework.} Building on the insights from \textbf{E1}, we propose the first FCGL framework, named POWER (gra\underline{P}h ev\underline{O}lution Trajectory-aware kno\underline{W}ledge Transf\underline{ER}). Specifically, POWER is designed with two fundamental objectives: (1) \underline{\textit{Addressing LGF at local clients}}: POWER selects experience nodes from past tasks with maximum local-global coverage and replays them during local training for future tasks; and (2) \underline{\textit{Tackling GEC at}} \underline{\textit{central server}}: POWER introduces a novel pseudo prototype reconstruction mechanism, enabling the server to capture multi-client graph evolution trajectories. Moreover, a trajectory-aware knowledge transfer process is applied to restore the lost knowledge of global GNN during parameter aggregation.

     \textbf{Our Contributions}: 
     (1) \textbf{Problem Identification.}
    To the best of our knowledge, this is the first exploration of the FCGL paradigm, offering a practical solution for continual graph learning under decentralized environments.
    (2) \textbf{In-depth Investigation.} (Sec.~\ref{sec: empirical investigation})
    We conduct in-depth empirical investigations of FCGL from the perspectives of \textcolor{darkred}{\textbf{Data}}, \textcolor{darkgreen}{\textbf{Feasibility}} and \textcolor{darkblue}{\textbf{Effectiveness}}, providing valuable insights for its development.
    (3) \textbf{Novel Framework.} (Sec.~\ref{sec: methodology})
    We propose POWER, an novel and effective FCGL training framework that tackles two non-trivial challenges of FCGL named LGF and GEC.
    (4) \textbf{State-of-the-art Performance.} (Sec.~\ref{sec: experiments})
   Experiments on eight datasets demonstrate that POWER consistently outperforms the federated extension of centralized CGL algorithms and vision-based federated continual learning algorithms.




    \section{Problem Formalization}
    For FCGL, there is a trusted central server and $K$ clients. Specifically, for the $k$-th client, we use $\mathcal{T}^k=\{T^k_1, T^k_2,...,T^k_M\}$ to denotes the set of its tasks, where $M$ is the number of tasks. Then, the local private evolving graph with $M$ tasks can be formalized as follows:
    \begin{equation}
    \begin{aligned}
    &\mathcal{G}^k=\{G^k_1, G^k_2, ..., G^k_M\};\\&G^k_{t}=G^k_{t-1}+\Delta G^k_t,
    \end{aligned}   
    \end{equation}
    \noindent where $G^k_t = (\mathcal{V}^k_t, \mathcal{E}^k_t)$ is the graph given in task $T^k$, Each node $v_i \in \mathcal{V}^k_t$ has a feature vector $\mathbf{x}_i \in \mathbb{R}^F$ and a one-hot label $\mathbf{y}_i \in \mathbb{R}^C$. The adjacency structure is represented by matrix $\mathbf{A}^k_t \in \mathbb{R}^{N_t \times N_t}$ ($N_t = |\mathcal{V}^k_t|$), and $\Delta G^k_t = (\Delta \mathcal{V}^k_t, \Delta \mathcal{E}^k_t) $ is the change of the node and edge set in task $T^k_t$. The $k$-th client aims to train models $\{\text{GNN}_{\mathbf{\Theta}_i^k}\}_{i=1}^t$
from streaming data, where $ \mathbf{\Theta}^k_i $ is the GNN parameter in task $ T^k_i $.

       This paper addresses semi-supervised node classification in a class-incremental setting, where new classes emerge as the graph evolves~\cite{de2021class_incre}. We assume all clients have $M$ tasks, each introducing $c$ new node categories and undergoing $R$ rounds of FCGL communication. Thus, after task $t = M$, each client encounters $C = c \times M$ categories. We adopt the widely-used FedAvg~\cite{mcmahan2017fedavg} aggregation strategy from federated learning, adapting it for the FCGL framework as follows: (1) \underline{\textit{Initialization.}} At the first communication round ($r\!=\!1$) of the first task ($t\!=\!1$), the central server sets the local GNN parameters of $K$ clients to the global GNN parameters, i.e., $\mathbf{\Theta}^{k} \leftarrow \mathbf{\Theta}^\text{g}\, \forall k$; (2) \underline{\textit{Local Updates.}} Each local GNN performs training on the current local data $G_t^k$ to minimize the task loss $\mathcal{L}(G_t^k; \mathbf{\Theta}_t^{k})$, and then updating the parameters: $\mathbf{\Theta}^k \leftarrow \mathbf{\Theta}^k - \eta \nabla \mathcal{L}$; (3) \underline{\textit{Global Aggregation.}} After local training, the server aggregates the local knowledge with respect to the number of training instances, i.e., $\mathbf{\Theta}^\text{g} \leftarrow \frac{N_k}{N} \sum_{k=1}^K \mathbf{\Theta}^k$ with $N = \sum_k N_k$, and distributes the global parameters $\mathbf{\Theta}^\text{g}$ to local clients selected at the next round. For each task $t$, the process alternates between steps 2 and 3 until reaching the final round $R$. The iteration continues until the last round of the last task.
       
\begin{figure*}[t]
  \includegraphics[width=0.998\textwidth]{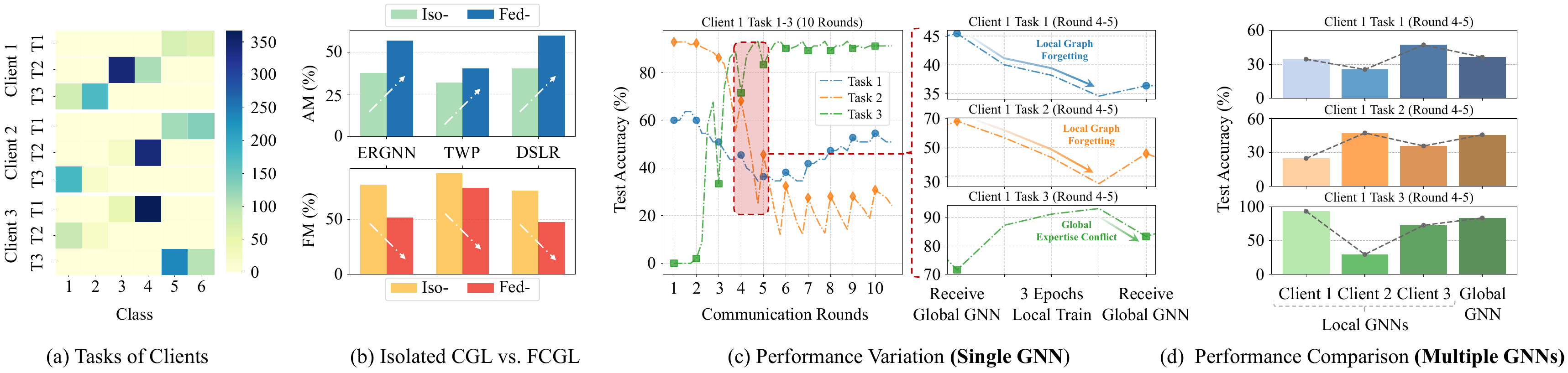}
  \captionsetup{font={small,stretch=1}}
  \caption{\textbf{Experimental results of our empirical study}. \textbf{(a) Node label distribution} of each client's class-incremental tasks, each exhibiting divergent evolution trajectories. \textbf{(b) Comparative analysis} of three CGL methods in isolated versus federated settings, presenting AM and FM metrics in the upper and lower parts, respectively. \textbf{(c) Performance variation} of Client 1's local GNN \textbf{(Single GNN)} during training on Client 1 Task 3, with markers denoting receiving global parameters from the server. \textbf{(d) Performance comparison} between clients and the server \textbf{(Multiple GNNs)} during training on Client 1 Task 3, indicating that the global GNN can be improved by local GNNs with expertise.}
\label{fig: empirical result}
\vspace{0.2cm}
\end{figure*}

\section{Empirical Investigation}
\label{sec: empirical investigation}

    In this section, we present a thorough empirical investigation of the proposed FCGL paradigm, structured around three key questions from different perspectives. From the \textcolor{darkred}{\textbf{Data}} perspective, to better understand the complexities of graph data in FCGL contexts, we answer \textcolor{darkred}{\textbf{Q1}}: What patterns and trends emerge in data distribution across multi-client evolving graphs in FCGL settings?  From the \textcolor{darkgreen}{\textbf{Feasibility}} perspective, to demonstrate the advantages of FCGL using collective intelligence, we answer \textcolor{darkgreen}{\textbf{Q2}}: Compared to isolated CGL, does a naive FCGL approach (i.e., CGL integrated with FedAvg) achieves better performance for decentralized evolving graphs? From the \textcolor{darkblue}{\textbf{Effectiveness}} perspective, to reveal the bottleneck of naive FCGL and realize efficient FCGL paradigm, we answer \textcolor{darkblue}{\textbf{Q3}}: What non-trivial challenges limit the performance of GNNs within the FCGL framework? Details for the described experimental setup, algorithms, and metrics refer to Sec.~\ref{sec: experimental setup}.

    To address \textcolor{darkred}{\textbf{Q1}}, we simulate multiple decentralized evolving graphs using a two-step approach. First, we partition the Cora dataset~\cite{Yang16cora} into three subgraphs using the Louvain algorithm~\cite{blondel2008louvain}, which is widely utilized in various FGL studies~\cite{li2024fedgta, zhang2021fedsage, zhang2024fgl_feddep}. These subgraphs are distributed among three different clients. Second, following the standard experimental settings of CGL~\cite{liu2021twp, zhou2021ergnn, choi2024dslr}, each client's local subgraph is further divided into three class-incremental tasks. Each task comprises two classes of nodes, discarding any surplus classes and removing inter-task edges while preserving only intra-task connections. For each client, we present its tasks with node label distributions in Fig.~\ref{fig: empirical result} (a). 
    
    \textcolor{darkred}{\textbf{Observation 1}}. Although all these clients eventually learn across six classes, they follow divergent \textit{graph evolution trajectories}, varying both in sample quantity (e.g., Clients 1 and 2) and in the sequence of class learning (e.g., Clients 1 and 3). These clients tend to develop expertise in specific classes at different stages, forming the basis of the collective intelligence harnessed by the FCGL paradigm.

    To address \textcolor{darkgreen}{\textbf{Q2}}, we assess the performance of three CGL algorithms, including ERGNN~\cite{zhou2021ergnn}, TWP~\cite{liu2021twp}, and DSLR~\cite{choi2024dslr}, under two scenarios: \textbf{isolated training} (referred to as Iso-ERGNN, Iso-TWP, and Iso-DSLR) and \textbf{federated training} using the FedAvg~\cite{mcmahan2017fedavg} algorithm (referred to as Fed-ERGNN, Fed-TWP, and Fed-DSLR). In the isolated setting, no client-server communication occurs, while the federated variants implement naive FCGL strategies that simply aggregate parameters from local GNNs. We assessed their performance using two metrics: the accuracy mean (AM) and the forgetting mean (FM), where \underline{higher} AM values denote better adaptation to new tasks, and \underline{lower} FM values denote less forgetting of previously learned tasks. The training protocol for each task includes 10 communication rounds, with 3 local epochs per round. As depicted in Fig.~\ref{fig: empirical result} (b), the federated variants consistently outperformed their isolated counterparts in both AM and FM metrics. This enhanced performance is attributed to FCGL's capability to leverage diverse knowledge from evolving graph data across clients, thereby reducing task forgetting through effective cross-client knowledge transfer (e.g., alleviating the knowledge forgetting of Client 1 Task 1 and Client 2 Task 1 through insights gained from Client 3 Task 3). 
    
    \textcolor{darkgreen}{\textbf{Observation 2}}. Even a basic implementation of FCGL can substantially enhance the performance of existing CGL algorithms on decentralized evolving graphs.

    To address \textcolor{darkblue}{\textbf{Q3}}, we evaluate the naive FCGL algorithm's performance across Client 1’s three tasks during the training of Task 3. For readability, we present the results obtained from Fed-ERGNN, as similar trends are also shown in Fed-TWP and Fed-DSLR. Our analysis is further concentrated on rounds 4 and 5. From the \textbf{Single GNN} aspect, we analyze performance variations in Client 1’s local GNN, particularly during the critical phases of \textit{global parameter reception} and \textit{local training}, motivated by findings in existing literature~\cite{yoon2021fedweit, liu2021twp}, which suggest that shifts in parameter values play a significant role in causing knowledge forgetting. This perspective enables a closer examination of how individual GNNs react to external updates. As shown in Fig.~\ref{fig: empirical result} (c), each \textit{global parameter reception} phase is marked with scatter points, while lines represent the 3-epoch \textit{local training} phases. From the \textbf{Multiple GNNs} aspect, we compare the performance of all local GNNs and the aggregated global GNN. The results are depicted in Fig.~\ref{fig: empirical result} (d).

    \begin{figure*}[htb]
      \includegraphics[width=0.998\textwidth]{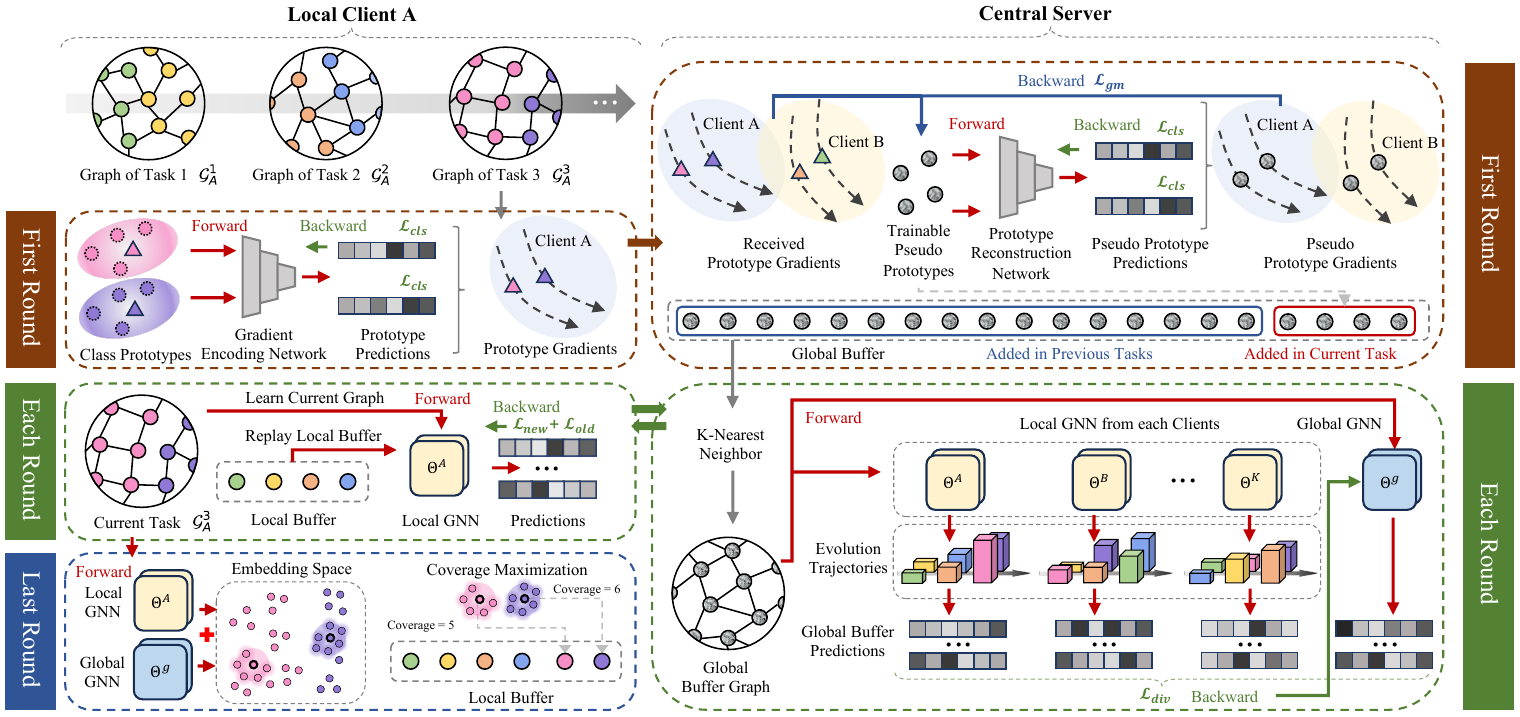}
      \captionsetup{font={small,stretch=1}}
      \caption{\textbf{Overview of our proposed POWER framework}. We follow a class-incremental setting, where each client collects an evolving graph into which unseen classes are continually introduced. Node colors indicate different labels. \textbf{(1) To tackle LGF}, POWER stores experience nodes with maximum local-global coverage in \textit{the last round of the old task} and replays them in \textit{each round of the new task}; \textbf{(2) To tackle GEC}, POWER reconstructs pseudo prototypes via client-server collaboration in \textit{the first round of each task}. Based on this, POWER constructs a global buffer graph and applies trajectory-aware knowledge transfer to recover the global GNN’s lost knowledge in \textit{each round of each task}.}
    \label{fig: framework}
    \end{figure*}

     \textcolor{darkblue}{\textbf{Observation 3}}. Building on these two aspects, we reveal the following two \textbf{Non-trivial Challenges} limit the performance of GNNs within the FCGL framework: (1) \underline{\textit{Local Graph Forgetting (LGF).}} From the \textbf{Single GNN} aspect, local training on the current task (i.e., Client 1 Task 3) significantly impairs performance on earlier tasks (i.e., Client 1 Tasks 1 and 2). This demonstrates that in the FCGL scenario, the local GNN \textbf{experiences local graph forgetting similar to that of centralized CGL}; and (2) \underline{\textit{Global Expertise Conflict (GEC).}} From the \textbf{Single GNN} aspect, incorporating global parameters can degrade the performance on the current task (i.e., Client 1 Task 3) while improving the performance on previous tasks (i.e., Client 1 Tasks 1 and 2). However, these improvements are limited when considering \textbf{Multiple GNNs} aspect. For example, in Task 1, the global GNN is outperformed by the local GNN of Client 3; in Task 2, it is outperformed by the local GNN of Client 2. This phenomenon occurs because each client's local GNN develops expertise shaped by its unique evolution trajectory, and this expertise leads to conflicts during parameter aggregation, where \textbf{the global GNN fails to capture it adequately}. Thus, GEC hinders FCGL's ability to adapt to new tasks and mitigate forgetting of previous ones.

    \textbf{In summary}, both \textbf{LGF} and \textbf{GEC} present non-trivial challenges that critically restrict the performance of GNNs in FCGL, which is crucial to be addressed in building an effective FCGL framework.

\section{POWER: The Proposed Framework}
\label{sec: methodology}

    In this section, we introduce POWER, the first FCGL framework designed for learning on multiple decentralized evolving graphs. We first provide an overview of POWER in Fig.~\ref{fig: framework}. Afterward, we delineate the architecture of POWER based on its two design objectives. Specifically, in Sec.~\ref{sec: power_lgf}, we detail the maximum local-global coverage-based experience node selection and replay mechanisms that POWER employs to \textit{address LGF at local clients}. In Sec.~\ref{sec: power_gec}, we expound on the pseudo prototype reconstruction mechanism and the graph evolution trajectory-aware knowledge transfer process, 
    which is employed to \textit{tackle GEC at central server}.

\subsection{Local-Global Coverage Maximization}
\label{sec: power_lgf}
    To tackle the challenge of LGF, POWER employs a strategic selection and storage of experience nodes from prior tasks, enabling targeted replay during local training on new tasks. We begin by introducing the experience node selection strategy.

\vspace{0.1cm}
\noindent\textbf{Experience Node Selection}. 
    To ensure that selected experience nodes capture knowledge from previous tasks, we build upon the Coverage Maximization (CM) strategy~\cite{zhou2021ergnn} with further modifications tailored to FCGL settings. The conventional CM is based on the intuitive assumption that nodes within the same category share similar properties in the embedding space, with many nodes of the same class naturally gathering near their class's representative nodes. As for FCGL, we can utilize the embeddings from both the local and global GNNs to fully consider both local and global knowledge. Specifically, consider the $k$-th client training on its $t$-th local task $T_t^k$; the local graph is denoted as $G_t^k$. At the last training round, we calculate the local-global node embeddings $\mathbf{Z}$ as follows:
    \begin{equation}
    \begin{aligned}
    \label{eq: local-global embedding}
        \mathbf{H} = \text{GNN}(&\mathbf{X}_t^k, \mathbf{A}_t^k|\mathbf{\Theta}^{k}), \\
        \mathbf{H}^\text{g} = \text{GNN}&(\mathbf{X}_t^k, \mathbf{A}_t^k| \mathbf{\Theta}^\text{g}),\\
        \mathbf{Z} = \alpha\mathbf{H}+&(1-\alpha)\mathbf{H}^\text{g},
    \end{aligned}
    \end{equation}
    \noindent where $\mathbf{H}$ and $\mathbf{H}^\text{g}$ are node embeddings from local and global GNNs with parameters $\mathbf{\Theta}^{k}$ and $\mathbf{\Theta}^\text{g}$, respectively. $\alpha$ is the trade-off factor for combining local and global embeddings. The local-global coverage  $C(v_i)$ for each node $v_i$ in the training set $\mathcal{V}^\text{lbl}$ can be calculated as:
    \begin{equation}
    \begin{aligned}
    \label{eq: local-global coverage}
        C(v_i) = | \{v_j \mid v_j \in \mathcal{V}^{\text{lbl}}, \mathbf{y}_i = \mathbf{y}_j, d(\mathbf{z}_i, \mathbf{z}_j) < \epsilon E(v_i)\}|,
    \end{aligned}
    \end{equation}
    \noindent
    \noindent where $d(\cdot,\cdot)$ measures the Euclidean distance between the embeddings of $v_i$ and $v_j$, $\epsilon$ serves as a threshold distance, and $E(v_i)$ denotes the average pairwise distance between the embeddings of training nodes in the same class as $v_i$ and the central node $v_i$.

    Subsequently, for each class $c$ in the current task, we select $b$ experience nodes from the set of nodes in class $c$ with maximum local-global coverage. The experience node set of class $c$ is denoted as $\mathcal{B}_{c}$, which is formulated as follows:
    \begin{equation}
    \begin{aligned}
    \label{eq: coverage max}
        \mathcal{B}_{c} = \argmax\nolimits_{\{v_{c_1},...,v_{c_b}| v_{c_1},...,v_{c_b} \in \mathcal{V}_c^\text{lbl}\}} \sum\nolimits_{i=1}^{b} C(v_{c_i}),
    \end{aligned}  
    \end{equation}
    \noindent where $\mathcal{V}_c^\text{lbl}=\{v_i | v_i \in \mathcal{V}^\text{lbl}, \mathbf{y}_i \!=\! c\}$, $b$ represents the buffer size allocated per task and class, constrained by the storage capacity budget. In our experiments, we set  $b=1$ as the default. Eq.~(\ref{eq: coverage max}) poses an NP-hard challenge, which we address using a greedy algorithm. At each iteration, the algorithm picks the node with the highest local-global coverage from the remaining unselected nodes in $\mathcal{V}_c^\text{lbl}$, repeating until $b$ nodes are selected.

    Finally, for each class $c$ in task $T_t^k$, the experience node set $ \mathcal{B}_c $ is stored into the local buffer $ \mathcal{B} $ for replay (i.e., $ \mathcal{B} = \mathcal{B} \cup \mathcal{B}_c $).

    \vspace{0.1cm}    
    \noindent\textbf{Experience Node Replay}. POWER's local training process pursues two primary objectives. First, to adapt to the current task $T_{t}^k $, the local GNN minimizes the cross-entropy loss over the labeled training set $ \mathcal{V}^\text{lbl} $, which can be calculated as follows:
    \begin{equation}
    \begin{aligned}
    \label{eq: new task loss}
        \hat{\mathbf{Y}} = \text{GNN-CLS}&(\mathbf{X}_{t}^k, \mathbf{A}_{t}^k \mid \mathbf{\Theta}^k), \\
        \mathcal{L}_{\text{new}} = -\!\!\sum\nolimits_{v_i \in \mathcal{V}^\text{lbl}} \mathbf{y}_i \log& \hat{\mathbf{y}}_i + (1 - \mathbf{y}_i) \log (1 - \hat{\mathbf{y}}_i),
    \end{aligned}  
    \end{equation}
    \noindent where $ \hat{\mathbf{Y}} $ denotes the soft labels predicted by the local GNN for the current task nodes, $ \mathbf{\Theta}^k $ represents the local GNN parameters, and $ \mathbf{y}_i $ is the ground-truth label for a specific training node $v_i$.

    When the current task is not the first task (i.e., $ t \neq 1 $), the client also optimizes a second objective, aiming at retaining knowledge from previously learned tasks. To achieve this, the client replays experience nodes stored in the local buffer $ \mathcal{B} $, which captures representative samples from prior tasks. The replay objective is the cross-entropy loss over these stored nodes, as formulated below:
    \begin{equation}
    \begin{aligned}
    \label{eq: old task loss}
       &\hat{\mathbf{Y}} = \text{GNN-CLS}(\mathbf{X}_\mathcal{B}, \mathbf{I} ~|~ \mathbf{\Theta}^k), \\
        \mathcal{L}_{\text{old}} = -\!&\sum\nolimits_{v_i\in\mathcal{B}}\mathbf{y}_i\log\hat{\mathbf{y}}_i + (1-\mathbf{y}_i)\log(1-\hat{\mathbf{y}}_i),
    \end{aligned}  
    \end{equation}
    \noindent where $\hat{\mathbf{Y}}$ represents the soft labels predicted by the local GNN for the buffer nodes, while $\mathbf{X}_\mathcal{B}$ denotes the feature matrix of nodes in the local buffer, and $\mathbf{y}_i$ is the ground truth for a specific buffer node $v_i$. Notably, since we store only the node features without including any topological structures, the adjacency matrix for buffer nodes is set as the identity matrix $ \mathbf{I} $, meaning that each buffer node is predicted in isolation.

    Finally, the loss function of local training is defined as:
    \begin{equation}
    \begin{aligned}
    \label{eq: local loss}
        \mathcal{L} = \beta\mathcal{L}_{\text{new}} + (1-\beta)\mathcal{L}_{\text{old}},
    \end{aligned}  
    \end{equation}
    \noindent where $\beta$ denotes the trade-off parameter to flexibly balance adaptation to new tasks with a replay of previous tasks.

\subsection{Trajectory-aware Knowledge Transfer}
\label{sec: power_gec}
    
    As discussed in Sec.~\ref{sec: empirical investigation}, GEC arises from knowledge conflicts during the aggregation of parameters from local GNNs trained on divergent graph evolution trajectories. Consequently, for certain classes, the global model exhibits sub-optimal performance compared to clients with expertise in those classes. Motivated by this, POWER employs two modules to address GEC: \textbf{(1)} \underline{\textit{Pseudo Prototype Reconstruction}}, which enables the server to discern each client’s evolution trajectory and expertise, and \textbf{(2)} \underline{\textit{Trajectory-aware Knowledge Transfer}}, which utilizes multi-client trajectories and expertise to recover the lost knowledge of the global GNN.

    \vspace{0.1cm}    
    \noindent \textbf{Pseudo Prototype Reconstruction Mechanism}. First, to capture the expertise of each client, POWER calculates the prototype (i.e., averaged node feature) for each class on the client side. Specifically, consider the $k$-th client on its first communication round of $t$-th local task $T_t^k$. We denote the labeled node set as $ \mathcal{V}^\text{lbl} $, with $ \mathcal{V}^\text{lbl}_c = \{ v_i \mid v_i \in \mathcal{V}^\text{lbl}, y_i = c \} $ representing the subset of nodes labeled as class $ c $.
    The prototype of class $c$ is denoted as $P_c$, formulated as:
    \begin{equation}
    \begin{aligned}
    \label{eq: prototype}
        P_c = \frac{1}{ |\mathcal{V}^\text{lbl}_c|} \sum\nolimits_{v_i \in \mathcal{V}^\text{lbl}_c} \mathbf{x}_i,
    \end{aligned}  
    \end{equation}
    However, directly transmitting class prototypes to the server poses significant privacy risks. To mitigate this, POWER adopts a privacy-preserving strategy where clients send prototype gradients instead, allowing the server to reconstruct pseudo prototypes via gradient matching. This enhances security for two reasons: (1) prototype gradients come from a randomly initialized network, making it hard to extract meaningful information without additional context, and (2) pseudo prototypes mimic the gradient patterns of true prototypes but lack identifiable numerical features that can be recognized by humans. Specifically, the client feeds the computed prototype $\mathbf{P}_c$ into an $L$-layer randomly initialized gradient encoding network $\Gamma$ with parameters $\{\mathbf{\Omega}_i\}_{i=1}^L$ to obtain predictions. We then backpropagate to compute the gradient $\nabla\Gamma_c$, whose $i$-th element $\nabla_{\mathbf{\Omega}_i}$ can be computed as follows:
    \begin{equation}
    \begin{aligned}
    \label{eq: prototype encode}
    \nabla_{\mathbf{\Omega}_i}\Gamma_c \!&=\! \nabla_{\mathbf{\Omega}_i}(-\mathbf{y}_c\log\hat{\mathbf{y}}_c\!+\!(1\!-\!\mathbf{y}_c)\log(1\!-\!\hat{\mathbf{y}}_c)),
    \end{aligned}  
    \end{equation}
    \noindent where $\hat{\mathbf{y}}_c = \Gamma(P_c | \{\mathbf{\Omega}_i\}_{i=1}^L)$ denotes the predicted soft label and $\mathbf{y}_c$ is the one-hot ground-truth vector representing class $c$.

    When the server receives the prototype gradient $\nabla\Gamma$, it reconstructs pseudo prototypes via gradient matching. Specifically, consider a trainable pseudo prototype $\hat{\mathbf{P}}$ initialized by a standard Gaussian noise $\mathcal{N}(0,1)$, the server obtains the pseudo prototype prediction by feeding  $\hat{\mathbf{P}}$ into a prototype reconstruction network $\Gamma$, which is the same as the gradient encoding network used by all local clients (i.e., also parameterized by $\{\mathbf{\Omega}_i\}_{i=1}^L$). Subsequently, the server backpropagates the cross-entropy loss to compute the pseudo prototype gradients $\nabla\hat{\Gamma}$, whose $i$-th element $\nabla_{\mathbf{\Omega}_i}$ is formulated as follows:
    \begin{equation}
    \begin{aligned}
    \label{eq: pseudo prototype encode}
    \nabla_{\mathbf{\Omega}_i}\hat{\Gamma} \!&=\! \nabla_{\mathbf{\Omega}_i}(-\mathbf{y}\log\hat{\mathbf{y}}\!+\!(1\!-\!\mathbf{y})\log(1\!-\!\hat{\mathbf{y}})),
    \end{aligned}  
    \end{equation}
    \noindent where $\hat{\mathbf{y}} = \Gamma(P | \{\mathbf{\Omega}_i\}_{i=1}^L)$ is the predicted soft label of pseudo prototype and $\mathbf{y}$ is the one-hot vector of a specific class $c$. Notably, the class $c$ of prototype gradient $\nabla\Gamma$ can be inferred with the symbol of prototype gradient (i.e., $c=\text{argmin}_{i}\nabla_{\Omega_{L_i}}\Gamma$). 
    
    Based on Eqs.~(\ref{eq: prototype encode}) and~(\ref{eq: pseudo prototype encode}), the trainable pseudo prototype is optimized via the gradient matching loss $\mathcal{L}_{\text{gm}}$, defined as:
    \begin{equation}
    \begin{aligned}
    \label{eq: gradient match loss}
        \mathcal{L}_{\text{gm}} = \sum\nolimits_{i=1}^L|\!| \nabla_{\mathbf{\Omega}_i}\Gamma - \nabla_{\mathbf{\Omega}_i}\hat{\Gamma}|\!|^2.
    \end{aligned}  
    \end{equation}
    \noindent Finally, the reconstructed pseudo prototypes are stored in global buffer $\mathcal{B}^\text{g}$ (i.e., $\mathcal{B}^\text{g}=\mathcal{B}^\text{g} \cup \{\hat{\mathbf{P}}\}$).

    Second, we define the evolution trajectory of each client through a cumulative label distribution, which aggregates task-specific label distributions across the client’s task sequence, applying a gradual decay to earlier tasks to simulate LGF during local training. Formally, the evolution trajectory for the $k$-th client on task $T^t$ is represented as $ \mathbf{q}_t^k $, defined as follows:
    \begin{equation}
    \begin{aligned}
    \label{eq: cumulative_label_distribution}
        \forall i \in \{1,...,t\}, \mathbf{p}_i^k &\propto \{ \sum\nolimits_{v_j \in \mathcal{V}^\text{lbl}}  \mathbf{1}_{y_j = c} \}_{c\in\mathcal{C}},\\
        \mathbf{q}_t^k =& \sum\nolimits_{i=1}^{t} \phi^{t-i}\mathbf{p}_i^k, 
    \end{aligned}  
    \end{equation}   
    \noindent where $\phi$ represents the decay coefficient that attenuates the contributions of past tasks, $\mathcal{C}$ denotes the total classes, $\mathbf{q}_i^k$ denotes the label distribution of task $T_i^k$, and $\mathbf{1}_{y_j = c}$ takes the value 1 if $y_j = c$ holds, and $0$ otherwise. Notably, Eq.~(\ref{eq: cumulative_label_distribution}) can be computed iteratively without storing nodes from previous tasks.

    \vspace{0.1cm}
    \noindent \textbf{Trajectory-aware Knowledge Transfer}. Our key insight is to utilize multi-client expertise and trajectories to recover the lost knowledge of the global GNN. First, based on the global buffer $\mathcal{B}^\text{g}$, the server constructs a global buffer graph $G^\text{g}$ via the K-Nearest Neighbor strategy to explore the global input space. Specifically, the feature matrix of the global buffer is denoted as $\mathbf{X}^\text{g}$, the constructed adjacency $\mathbf{A}^\text{g}$ can be calculated as:
    \begin{equation}
    \begin{aligned}
    \label{eq: knn}
    &\mathbf{H}\!=\! \sigma(\mathbf{X}^\text{g}\mathbf{X}^{\text{g}^T}\!),\\
    \mathbf{A}^\text{g}[u,v]&\!=\!\left\{
    \begin{aligned}
    &1,   \text{if }  v \in \text{TOPK}(\bar{k}, \mathbf{H}[u,:]), \\
    &0,  \text{otherwise},
    \end{aligned}
    \right.
    \end{aligned}
    \end{equation}
    \noindent where $\sigma(\cdot)$ represents the element-wise sigmoid function, and $\text{TOPK}(\bar{k}, \cdot)$ selects the index of the first $\bar{k}$ largest items.

    Subsequently, POWER leverages the global buffer graph $ \mathcal{G}^{\text{g}} $ along with the set of clients' evolution trajectories $ \mathbf{q}_t = \{\mathbf{q}_t^k\}_{k=1}^K$ to facilitate knowledge transfer. The extent to the global model’s integration of expertise from each local GNN is guided by the client’s evolution trajectory. Specifically, the knowledge transfer loss for training on task $ T_t $ is defined as:
    \begin{align}
    \label{eq: div loss}
    \forall k \in \{1, ..&., K\}, \hat{\mathbf{Y}}^{k} = \text{GNN-CLS}(\mathbf{X}^\text{g}, \mathbf{A}^\text{g}|\mathbf{\Theta}^k),  \nonumber\\
    \hat{\mathbf{Y}}^\text{g} &= \text{GNN-CLS}(\mathbf{X}^\text{g}, \mathbf{A}^\text{g}|\mathbf{\Theta}^\text{g});  \\
    \mathcal{L}_\text{div} = \sum_{c=1}^C\sum_{k=1}^K& \sum_{v_i\in \mathcal{V}^\text{g}}  \mathbf{1}(\mathbf{y}_i=c)\frac{\mathbf{q}_t^k(c)}{\sum\nolimits_{j=1}^K\mathbf{q}_t^j(c)}\text{KL}(\hat{\mathbf{y}}_i^\text{g}|\!|\hat{\mathbf{y}}_i^{k}), \nonumber
    \end{align}
    \noindent where $ \mathcal{C} $ is the set of all classes, $ \mathcal{V}^\text{g}$ represents the node set of the global buffer, $\mathbf{y}_i$ denotes the ground truth for the $i$-th node in the global buffer graph, and $\hat{\mathbf{y}}_i^g$ and $\hat{\mathbf{y}}_i^k$ are the soft labels predicted by the global GNN and the local GNN of the $k$-th client, respectively. The term $\frac{\mathbf{q}t^k(c)}{\sum{j=1}^K \mathbf{q}_t^j(c)}$ provides the proportion of class $ c $ within the evolution trajectory of the $k$-th client relative to all clients, and $ \text{KL}(\cdot |\!| \cdot) $ denotes the Kullback-Leibler divergence.
    The full algorithm for POWER is outlined in Algorithm~\ref{alg: power-all}.

   \begin{algorithm}[htbp]\fontsize{8pt}{4pt}\selectfont
    \DontPrintSemicolon
    \SetAlgoLined
    \caption{POWER-Clients and Server Execution}
    \label{alg: power-all}
    \SetKwFunction{FServer}{S\_Exec}
    \SetKwFunction{FClient}{C\_Exec}
    \SetKwProg{Fn}{Function}{:}{end}

    \KwInput{Number of tasks $T$, each task communication rounds $R$, local epochs $E$, global epochs $E_g$, global GNN parameter $\mathbf{\Theta}^\text{g}$}
    \KwOutput{Local GNN parameters $\{\mathbf{\Theta}^k\}_{k=1}^K$}
    
    \For{each training task $t$ \textup{in} $1, ..., T$}{
        \For{each communication round $r$ \textup{in} $1,...,R$}{
            \For{each client $k=1,...,K$}{
                $\mathbf{\Theta}^k$, $\nabla\Gamma_t^k$, $\mathbf{q}_t^k \gets $\FClient{$k, t,r,\mathbf{\Theta}^\text{g}$}
            }
            $\mathbf{\Theta}^\text{g} \gets$\FServer{$t$, $r$, $\{\nabla\Gamma_t^k\}_{k=1}^{K}$, $\{\mathbf{q}_t^k\}_{k=1}^K$, $\{\mathbf{\Theta}^k\}_{k=1}^K$}\tcp*{Receive uploaded messages}
        }
    }

    \tcc{Client Execution}
    \Fn{\FClient{$k, t,r,\mathbf{\Theta}^\text{g}$}}{
            {$\mathbf{\Theta}^\text{k} \gets \mathbf{\Theta}^\text{g}$}\tcp*{Update local GNN parameters}
            \For{each local training epoch $e$ \textup{in} $1, ..., E$}{
                Perform local training and replaying.\tcp*{Eq.~(\ref{eq: local loss})}
            }
            \If{$r = R$}{ 
                \For{each class $c$ \textup{in} task $T_t^k$}{
                Select experience nodes $\mathcal{B}_c$\tcp*{Eqs.~(\ref{eq: local-global embedding}), (\ref{eq: local-global coverage}), (\ref{eq: coverage max})}

                $\mathcal{B}\gets\mathcal{B}\cup\mathcal{B}_c$\tcp*{Store to local buffer}
                }
            }
            \eIf{$r = 1$}{
                $\nabla\Gamma_t^k \gets \{\}$
                
                \For{each class $c$ \textup{in} task $T_t^k$}{
                    Compute prototype gradients $\nabla\Gamma_c$\tcp*{Eq.~(\ref{eq: prototype encode})}
                    $\nabla\Gamma_t^k \gets \nabla\Gamma_k^t \cup \nabla\Gamma_c $
                }
                
                Compute evolution trajectory $\mathbf{q}_t^k$\tcp*{Eq.~(\ref{eq: cumulative_label_distribution})}
                \Return{$\mathbf{\Theta}^k$, $\nabla\Gamma_t^k$, $\mathbf{q}_t^k$}
            }
            {
                \Return{$\mathbf{\Theta}^k$, None, None}
            }
    }
    
    \tcc{Server Execution}
    
    \Fn{\FServer{$t$, $r$, $\nabla\Gamma_t$, $\mathbf{q}_t$, $\mathbf{\Theta}$}}{
        \If{$r = 1$}{
                \For{$\nabla \Gamma_t^k$ \textup{in} $\nabla \Gamma_t$}{
                    \For{$\nabla \Gamma_{t_c}^k$ \textup{in} $\nabla \Gamma_t^k$}{
                         Reconstruct $\hat{\mathbf{P}}$ via $\nabla \Gamma_{t_c}^k$\tcp*{Eqs.~(\ref{eq: pseudo prototype encode}), (\ref{eq: gradient match loss})}
                        
                        \tcc{Store to global buffer}
                         $\mathcal{B}^\text{g} \gets \mathcal{B}^\text{g} \cup \{\hat{\mathbf{P}}\}$
                    }
                }
            }
            \For{each global training epoch $e$ \textup{in} $1, ..., E_g$}{
                Construct global buffer graph $G^\text{g}$\tcp*{Eq.~(\ref{eq: knn})}
                Perform Knowledge transfer\tcp*{Eq.~(\ref{eq: div loss})}

            }
        \Return{Global GNN parameters $\mathbf{\Theta}^\text{g}$}\;
    }
    \end{algorithm}

\section{Experiments}
\label{sec: experiments}

    In this section, we present a comprehensive evaluation of POWER. We first introduce the settings of our experiments (Sec.~\ref{sec: experimental setup}). 
    After that, we aim to address the following questions:
    \textbf{Q1}: Compared with state-of-the-art baselines, can POWER achieve better predictive performance under FCGL scenario (Sec.~\ref{sec: effectiveness})?
    \textbf{Q2}: How much does each module in POWER contribute to the overall performance (Sec.~\ref{sec: ablation study})? 
    \textbf{Q3}: Can POWER remain overall robust under changes in hyperparameters (Sec.~\ref{sec: sensitivity})? 
    Moreover, we also provide further experimental investigations about the efficiency (\textbf{Q4}) and sparsity robustness (\textbf{Q5}) of POWER in~\cite{zhu2024power_tech}~\ref{appendix: efficiency} and~\ref{appendix: sparsity robustness}, respectively.

\begin{table*}[t]
\setlength{\abovecaptionskip}{0.25cm}
\renewcommand{\arraystretch}{1.5}
\caption{\textbf{Statistical information of the experimental datasets}, where ``\#Classes pT'' denotes the number of classes per task.
}
\footnotesize 
\label{tab: datasets}
\resizebox{\linewidth}{25mm}{
\setlength{\tabcolsep}{2mm}{
\begin{tabular}{cccccccccc}

\toprule[1pt]

Datasets        & \#Nodes       & \#Features    & \#Edges       & \#Classes    & \#Classes pT & \#Tasks & \#Clients & Train/Val/Test        & Description               \\ 
\midrule[0.1pt]
Cora            & 2,708         & 1,433         & 5,429         & 7            & 2          & 3         & 3         & 20\%/40\%/40\%        & Citation Network          \\
CiteSeer        & 3,327         & 3,703         & 4,732         & 6            & 2          & 3         & 3         & 20\%/40\%/40\%        & Citation Network          \\
OGB-arxiv       & 169,343       & 128           & 231,559       & 40           & 10          & 4        & 5         & 60\%/20\%/20\%        &  Citation Network         \\
Computers       & 13,381        & 767           & 245,778       & 10           & 2          & 5         & 3         & 20\%/40\%/40\%        & Co-purchase Network       \\
Physics         & 34,493        & 8,415         & 247,962       & 5            & 2          & 2         & 10        & 20\%/40\%/40\%        & Co-authorship Network     \\ 
Squirrel        & 5,201         & 2,089         & 216,933       & 5            & 2          & 2         & 10        & 48\%/32\%/20\%        & Wiki-page Network         \\
Flickr          & 89,250        & 500           & 899,756       & 7            & 2          & 3         & 10        &  60\%/20\%/20\%       & Image Network             \\ 
Roman-empire    & 22,662        & 300           & 32,927        & 18           & 6          & 3         & 5         & 50\%/25\%/25\%        & Article Syntax Network    \\

\bottomrule[1pt]

\end{tabular}
}}
\vspace{0.1cm}
\end{table*}

\subsection{Experimental Setup}
\label{sec: experimental setup}

    We introduce 8 benchmark graph datasets across 6 domains and the FCGL simulation strategy, followed by 10 state-of-the-art baselines and detailed metrics descriptions. For further statistics and experimental details, see~\cite{zhu2024power_tech}~\ref{appendix: exp setting}.

\vspace{0.18cm}
\noindent \textbf{Datasets and Simulation Strategy.}
    We evaluate POWER on 8 benchmark graph datasets across 6 domains, including citation networks (Cora, Citeseer~\cite{Yang16cora}, OGB-arxiv~\cite{hu2020ogb}), co-purchase (Computers~\cite{shchur2018amazon_datasets}), co-authorship (Physics~\cite{shchur2018amazon_datasets}), wiki-page (Squirrel~\cite{pei2020geomgcn}), image (Flickr~\cite{zeng2019graphsaint}), and article syntax networks (Roman-empire~\cite{platonov2023hete_gnn_survey4}). The decentralized evolving graph simulation strategy is detailed in Sec.~\ref{sec: empirical investigation}. For each dataset, client counts, the class counts of each task, and data splits are defined based on node and class counts of the original dataset.

\vspace{0.18cm}
\noindent \textbf{Simulation Details.}
As mentioned in Sec.~\ref{sec: empirical investigation} and Sec.~\ref{sec: experimental setup}, the decentralized evolving graph simulation strategy is a 2-step process, which first uses the Louvain~\cite{blondel2008louvain} algorithm to partition the original graph dataset into multiple subgraphs. These subgraphs are subsequently distributed to different clients, all of which are further divided into multiple class-incremental tasks. Each task comprises multiple classes of nodes, discarding any surplus classes and removing inter-task edges while preserving intra-task connections. For each dataset, the number of clients, class per task, and data splits are defined based on the node and class counts of the original graph, as provided in Table.~\ref{tab: datasets}.

\vspace{0.18cm}
\noindent \textbf{Baselines.}
    We summarize 10 baselines into 3 categories as follows: \textbf{(1) FL/FGL Fine-tuning}, including FedAvg~\cite{mcmahan2017fedavg}, FedSage+~\cite{zhang2021fedsage}, Fed-PUB~\cite{baek2022fedpub} and FedGTA~\cite{li2024fedgta}, which simply replace the local client training process of the FL/FGL algorithm with fine-tuning on sequential tasks; \textbf{(2) Federated CGL}, including Fed-ERGNN, FedTWP and Fed-DSLR, which combines the naive FedAvg algorithm with three representative CGL methods, ERGNN~\cite{zhou2021ergnn}, TWP~\cite{liu2021twp} and DSLR~\cite{choi2024dslr}, respectively; \textbf{(3) Federated Continual Learning for CV}, which includes GLFC~\cite{dong2022glfc}, TARGET~\cite{zhang2023target} and LANDER~\cite{tran2024lander}.
    
\vspace{0.18cm}
\noindent \textbf{Evaluation Metrics.} We adopt two primary metrics commonly used in continual learning benchmarks~\cite{zhou2021ergnn, liu2021twp, choi2024dslr} to evaluate model performance across sequential tasks: \textbf{(1) Accuracy Mean} (AM), defined as 
$\text{AM} = \frac{1}{T}\sum_{i=1}^T \mathbf{A}_{T,i},$
and \textbf{(2) Forgetting Mean} (FM), given by $\text{FM} = \frac{1}{T-1}\sum_{i=1}^{T-1} (A_{i,i} - A_{T,i}),$
where $ T $ represents the total number of tasks. In standard CGL, $ \mathbf{A}_{i,j} $ denotes the accuracy on task $ j $ following the completion of task $ i $. In our FCGL framework, each client’s local GNN predicts its own task, and we compute a weighted average of \( \mathbf{A}_{i,j} \) across clients, based on their sample counts. AM measures the average accuracy on each task after training on the final task, with higher values indicating stronger performance. FM captures the mean performance drop across tasks as new ones are learned, with lower values indicating better retention.

\begin{table*}[htbp]
    \setlength{\abovecaptionskip}{0.2cm}
    \renewcommand{\arraystretch}{2.2}
    \caption{\textbf{Performance comparison of POWER and baselines}, where the best and second results are highlighted in \textbf{bold} and \underline{underline}.}
    \footnotesize 
    \label{tab: compare baseline}
    \resizebox{\linewidth}{48mm}{
    \setlength{\tabcolsep}{0mm}{
    \begin{tabular}{c|cc|cc|cc|cc|cc|cc|cc|cc}
        \toprule[1pt]
         Dataset & \multicolumn{2}{c|}{Cora} & \multicolumn{2}{c|}{CiteSeer} & \multicolumn{2}{c|}{OGB-arxiv} & \multicolumn{2}{c|}{Computers} & \multicolumn{2}{c|}{Physics} & \multicolumn{2}{c|}{Squirrel} & \multicolumn{2}{c|}{Flickr} & \multicolumn{2}{c}{Roman-empire}\\
         
        \midrule[0.1pt]    
        
        Methods & AM $\uparrow$ & FM $\downarrow$ & AM $\uparrow$ & FM $\downarrow$ & AM $\uparrow$ & FM $\downarrow$ & AM $\uparrow$ & FM $\downarrow$ & AM $\uparrow$ & FM $\downarrow$ & AM $\uparrow$ & FM $\downarrow$ & AM $\uparrow$ & FM $\downarrow$ & AM $\uparrow$ & FM $\downarrow$ \\
        
        \midrule[0.1pt]
        FedAvg~\cite{mcmahan2017fedavg}
        & \parbox[c]{1cm}{\centering 40.14 \\ \tiny $\pm$1.01}
        & \parbox[c]{1cm}{\centering 78.92 \\ \tiny $\pm$1.84} 
        & \parbox[c]{1cm}{\centering 42.50 \\ \tiny $\pm$0.73}
        & \parbox[c]{1cm}{\centering 59.30 \\ \tiny $\pm$1.11}
        & \parbox[c]{1cm}{\centering 21.90 \\ \tiny $\pm$0.99}
        & \parbox[c]{1cm}{\centering 57.57 \\ \tiny $\pm$1.54}
        & \parbox[c]{1cm}{\centering 19.81 \\ \tiny $\pm$0.07}
        & \parbox[c]{1cm}{\centering 94.94 \\ \tiny $\pm$1.71}
        & \parbox[c]{1cm}{\centering 60.63 \\ \tiny $\pm$1.80}
        & \parbox[c]{1cm}{\centering 75.98 \\ \tiny $\pm$3.58}
        & \parbox[c]{1cm}{\centering 32.33 \\ \tiny $\pm$0.95}
        & \parbox[c]{1cm}{\centering 59.03 \\ \tiny $\pm$2.67}
        & \parbox[c]{1cm}{\centering 27.46 \\ \tiny $\pm$2.00}
        & \parbox[c]{1cm}{\centering 37.35 \\ \tiny $\pm$3.02}
        & \parbox[c]{1cm}{\centering 17.19 \\ \tiny $\pm$0.92}
        & \parbox[c]{1cm}{\centering 42.11 \\ \tiny $\pm$1.60} \\
        
        FedSage+~\cite{zhang2021fedsage}        
        & \parbox[c]{1cm}{\centering 30.41 \\ \tiny $\pm$2.39}
        & \parbox[c]{1cm}{\centering 87.63 \\ \tiny $\pm$0.28}
        & \parbox[c]{1cm}{\centering 33.70 \\ \tiny $\pm$4.51}
        & \parbox[c]{1cm}{\centering 58.76 \\ \tiny $\pm$6.29}
        & \parbox[c]{1cm}{\centering 15.22 \\ \tiny $\pm$0.83} 
        & \parbox[c]{1cm}{\centering 56.87 \\ \tiny $\pm$1.21} 
        & \parbox[c]{1cm}{\centering 12.22 \\ \tiny $\pm$1.77} 
        & \parbox[c]{1cm}{\centering 95.19 \\ \tiny $\pm$2.35}  
        & \parbox[c]{1cm}{\centering 58.24 \\ \tiny $\pm$2.18} 
        & \parbox[c]{1cm}{\centering 79.82 \\ \tiny $\pm$3.04}  
        & \parbox[c]{1cm}{\centering 29.52 \\ \tiny $\pm$2.18} 
        & \parbox[c]{1cm}{\centering 79.82 \\ \tiny $\pm$3.04}  
        & \parbox[c]{1cm}{\centering 25.58 \\ \tiny $\pm$2.54} 
        & \parbox[c]{1cm}{\centering 36.12 \\ \tiny $\pm$2.07}
        & \parbox[c]{1cm}{\centering 13.35 \\ \tiny $\pm$1.33} 
        & \parbox[c]{1cm}{\centering 40.27 \\ \tiny $\pm$3.15} \\

        Fed-PUB~\cite{baek2022fedpub}       
        & \parbox[c]{1cm}{\centering 30.02 \\ \tiny $\pm$1.28}
        & \parbox[c]{1cm}{\centering 90.88 \\ \tiny $\pm$0.35}
        & \parbox[c]{1cm}{\centering 33.48 \\ \tiny $\pm$2.34}
        & \parbox[c]{1cm}{\centering 73.66 \\ \tiny $\pm$2.38}
        & \parbox[c]{1cm}{\centering 15.17 \\ \tiny $\pm$0.15}
        & \parbox[c]{1cm}{\centering 56.94 \\ \tiny $\pm$0.74}
        & \parbox[c]{1cm}{\centering 11.93 \\ \tiny $\pm$1.54}
        & \parbox[c]{1cm}{\centering 95.46 \\ \tiny $\pm$0.46}
        & \parbox[c]{1cm}{\centering 58.75 \\ \tiny $\pm$0.62}
        & \parbox[c]{1cm}{\centering 79.72 \\ \tiny $\pm$1.24}
        & \parbox[c]{1cm}{\centering 28.82 \\ \tiny $\pm$0.42}
        & \parbox[c]{1cm}{\centering 43.52 \\ \tiny $\pm$0.56}
        & \parbox[c]{1cm}{\centering 25.47 \\ \tiny $\pm$1.10}
        & \parbox[c]{1cm}{\centering 36.27 \\ \tiny $\pm$1.27}
        & \parbox[c]{1cm}{\centering 11.56 \\ \tiny $\pm$0.20}
        & \parbox[c]{1cm}{\centering 41.60 \\ \tiny $\pm$1.02} \\

        FedGTA~\cite{li2024fedgta}        
        & \parbox[c]{1cm}{\centering 38.62 \\ \tiny $\pm$0.66}
        & \parbox[c]{1cm}{\centering 80.99 \\ \tiny $\pm$1.31}
        & \parbox[c]{1cm}{\centering 47.00 \\ \tiny $\pm$0.84}
        & \parbox[c]{1cm}{\centering 54.05 \\ \tiny $\pm$1.65}
        & \parbox[c]{1cm}{\centering 16.80 \\ \tiny $\pm$0.62}
        & \parbox[c]{1cm}{\centering 62.89 \\ \tiny $\pm$1.83}
        & \parbox[c]{1cm}{\centering 19.48 \\ \tiny $\pm$0.60}
        & \parbox[c]{1cm}{\centering 96.40 \\ \tiny $\pm$3.35}
        & \parbox[c]{1cm}{\centering 61.62 \\ \tiny $\pm$2.07}
        & \parbox[c]{1cm}{\centering 74.10 \\ \tiny $\pm$4.09}
        & \parbox[c]{1cm}{\centering 30.84 \\ \tiny $\pm$0.38}
        & \parbox[c]{1cm}{\centering 42.47 \\ \tiny $\pm$2.29}
        & \parbox[c]{1cm}{\centering 27.78 \\ \tiny $\pm$1.82}
        & \parbox[c]{1cm}{\centering 40.05 \\ \tiny $\pm$1.74}
        & \parbox[c]{1cm}{\centering 14.16 \\ \tiny $\pm$1.29}
        & \parbox[c]{1cm}{\centering 46.54 \\ \tiny $\pm$0.66} \\
        
        \midrule[0.1pt]

        Fed-TWP~\cite{liu2021twp}
        & \parbox[c]{1cm}{\centering 40.34 \\ \tiny $\pm$1.10}
        & \parbox[c]{1cm}{\centering 78.67 \\ \tiny $\pm$1.63}
        & \parbox[c]{1cm}{\centering 42.38 \\ \tiny $\pm$1.33}
        & \parbox[c]{1cm}{\centering 58.94 \\ \tiny $\pm$1.48}
        & \parbox[c]{1cm}{\centering 22.44 \\ \tiny $\pm$0.35}
        & \parbox[c]{1cm}{\centering 57.31 \\ \tiny $\pm$1.42}
        & \parbox[c]{1cm}{\centering 19.02 \\ \tiny $\pm$1.81}
        & \parbox[c]{1cm}{\centering 94.11 \\ \tiny $\pm$3.65}
        & \parbox[c]{1cm}{\centering 61.43 \\ \tiny $\pm$2.22}
        & \parbox[c]{1cm}{\centering 74.38 \\ \tiny $\pm$4.45}
        & \parbox[c]{1cm}{\centering 32.34 \\ \tiny $\pm$0.55}
        & \parbox[c]{1cm}{\centering 57.08 \\ \tiny $\pm$3.84}
        & \parbox[c]{1cm}{\centering 30.43 \\ \tiny $\pm$4.20}
        & \parbox[c]{1cm}{\centering 37.41 \\ \tiny $\pm$6.37}
        & \parbox[c]{1cm}{\centering 17.58 \\ \tiny $\pm$0.86}
        & \parbox[c]{1cm}{\centering 42.02 \\ \tiny $\pm$1.36} \\

        Fed-ERGNN~\cite{zhou2021ergnn} 
        & \parbox[c]{1cm}{\centering 60.40 \\ \tiny $\pm$2.78}
        & \parbox[c]{1cm}{\centering 45.68 \\ \tiny $\pm$4.20}
        & \parbox[c]{1cm}{\centering \underline{49.92} \\ \tiny $\pm$2.02}
        & \parbox[c]{1cm}{\centering \underline{39.19} \\ \tiny $\pm$2.41} 
        & \parbox[c]{1cm}{\centering \underline{32.63} \\ \tiny $\pm$3.48}
        & \parbox[c]{1cm}{\centering \underline{48.78} \\ \tiny $\pm$5.75}
        & \parbox[c]{1cm}{\centering 39.79 \\ \tiny $\pm$7.00}
        & \parbox[c]{1cm}{\centering 20.27 \\ \tiny $\pm$13.82}
        & \parbox[c]{1cm}{\centering \underline{85.23} \\ \tiny $\pm$2.74}
        & \parbox[c]{1cm}{\centering 26.02 \\ \tiny $\pm$5.56} 
        & \parbox[c]{1cm}{\centering 35.13 \\ \tiny $\pm$2.12}
        & \parbox[c]{1cm}{\centering 41.85 \\ \tiny $\pm$4.38}
        & \parbox[c]{1cm}{\centering \underline{31.79} \\ \tiny $\pm$3.93}
        & \parbox[c]{1cm}{\centering \underline{22.52} \\ \tiny $\pm$5.85}
        & \parbox[c]{1cm}{\centering \underline{24.71} \\ \tiny $\pm$0.87}
        & \parbox[c]{1cm}{\centering \underline{27.87} \\ \tiny $\pm$3.56}\\

        Fed-DSLR~\cite{choi2024dslr}
        & \parbox[c]{1cm}{\centering \underline{61.21} \\ \tiny $\pm$3.25}
        & \parbox[c]{1cm}{\centering \underline{43.77} \\ \tiny $\pm$5.10}
        & \parbox[c]{1cm}{\centering 49.55 \\ \tiny $\pm$2.16}
        & \parbox[c]{1cm}{\centering 40.92 \\ \tiny $\pm$3.27} 
        & \parbox[c]{1cm}{\centering 31.69 \\ \tiny $\pm$4.32}
        & \parbox[c]{1cm}{\centering 50.32 \\ \tiny $\pm$3.22}
        & \parbox[c]{1cm}{\centering \underline{40.32} \\ \tiny $\pm$5.30}
        & \parbox[c]{1cm}{\centering \underline{18.56} \\ \tiny $\pm$8.31}
        & \parbox[c]{1cm}{\centering 84.75 \\ \tiny $\pm$5.74}
        & \parbox[c]{1cm}{\centering 28.12 \\ \tiny $\pm$6.56} 
        & \parbox[c]{1cm}{\centering 35.21 \\ \tiny $\pm$3.15}
        & \parbox[c]{1cm}{\centering 41.33 \\ \tiny $\pm$5.27}
        & \parbox[c]{1cm}{\centering 31.55 \\ \tiny $\pm$2.47}
        & \parbox[c]{1cm}{\centering 24.18 \\ \tiny $\pm$6.30}
        & \parbox[c]{1cm}{\centering 24.32 \\ \tiny $\pm$2.87}
        & \parbox[c]{1cm}{\centering 28.22 \\ \tiny $\pm$4.13}\\

        \midrule[0.1pt]

        GLFC~\cite{dong2022glfc}        

        & \parbox[c]{1cm}{\centering 52.04 \\ \tiny $\pm$5.34}
        & \parbox[c]{1cm}{\centering 55.80 \\ \tiny $\pm$7.13}
        & \parbox[c]{1cm}{\centering 46.30 \\ \tiny $\pm$3.27} 
        & \parbox[c]{1cm}{\centering 51.24\\ \tiny $\pm$4.35} 
        & \parbox[c]{1cm}{\centering 30.27 \\ \tiny $\pm$4.34}
        & \parbox[c]{1cm}{\centering 52.43 \\ \tiny $\pm$5.25}
        & \parbox[c]{1cm}{\centering 38.43 \\ \tiny $\pm$7.62}
        & \parbox[c]{1cm}{\centering 23.56 \\ \tiny $\pm$3.27}
        & \parbox[c]{1cm}{\centering 67.01 \\ \tiny $\pm$4.87}
        & \parbox[c]{1cm}{\centering 60.97 \\ \tiny $\pm$4.22}
        & \parbox[c]{1cm}{\centering \underline{35.99} \\ \tiny $\pm$6.78}
        & \parbox[c]{1cm}{\centering \underline{40.21} \\ \tiny $\pm$4.28}
        & \parbox[c]{1cm}{\centering 30.80 \\ \tiny $\pm$3.74}
        & \parbox[c]{1cm}{\centering 35.27 \\ \tiny $\pm$6.03}
        & \parbox[c]{1cm}{\centering 16.86 \\ \tiny $\pm$3.07}
        & \parbox[c]{1cm}{\centering 41.52 \\ \tiny $\pm$4.85} \\
        
        TARGET~\cite{zhang2023target}         
        & \parbox[c]{1cm}{\centering 51.64 \\ \tiny $\pm$4.18}
        & \parbox[c]{1cm}{\centering 56.69 \\ \tiny $\pm$2.37}
        & \parbox[c]{1cm}{\centering 44.62 \\ \tiny $\pm$2.14}
        & \parbox[c]{1cm}{\centering 56.95 \\ \tiny $\pm$3.15} 
        & \parbox[c]{1cm}{\centering 30.52 \\ \tiny $\pm$5.19} 
        & \parbox[c]{1cm}{\centering 53.02 \\ \tiny $\pm$4.20} 
        & \parbox[c]{1cm}{\centering 37.75 \\ \tiny $\pm$1.54}
        & \parbox[c]{1cm}{\centering 26.82 \\ \tiny $\pm$1.79}
        & \parbox[c]{1cm}{\centering 75.48 \\ \tiny $\pm$2.06}
        & \parbox[c]{1cm}{\centering \underline{19.88} \\ \tiny $\pm$3.18}
        & \parbox[c]{1cm}{\centering 32.75 \\ \tiny $\pm$2.03}
        & \parbox[c]{1cm}{\centering 51.59 \\ \tiny $\pm$3.11} 
        & \parbox[c]{1cm}{\centering 31.07 \\ \tiny $\pm$3.24}
        & \parbox[c]{1cm}{\centering 25.12 \\ \tiny $\pm$2.20}
        & \parbox[c]{1cm}{\centering 14.37 \\ \tiny $\pm$1.11}
        & \parbox[c]{1cm}{\centering 45.02 \\ \tiny $\pm$1.95} \\
        
        LANDER~\cite{tran2024lander}        
        & \parbox[c]{1cm}{\centering 52.85 \\ \tiny $\pm$5.07}
        & \parbox[c]{1cm}{\centering 53.57 \\ \tiny $\pm$3.24}
        & \parbox[c]{1cm}{\centering 45.93 \\ \tiny $\pm$3.20}
        & \parbox[c]{1cm}{\centering 52.45 \\ \tiny $\pm$4.33} 
        & \parbox[c]{1cm}{\centering 30.45 \\ \tiny $\pm$4.12} 
        & \parbox[c]{1cm}{\centering 54.31 \\ \tiny $\pm$3.73} 
        & \parbox[c]{1cm}{\centering 38.13 \\ \tiny $\pm$2.64}
        & \parbox[c]{1cm}{\centering 24.27 \\ \tiny $\pm$3.28}
        & \parbox[c]{1cm}{\centering 73.24 \\ \tiny $\pm$4.51}
        & \parbox[c]{1cm}{\centering 22.46 \\ \tiny $\pm$4.20}
        & \parbox[c]{1cm}{\centering 33.21 \\ \tiny $\pm$3.35}
        & \parbox[c]{1cm}{\centering 52.03 \\ \tiny $\pm$5.17} 
        & \parbox[c]{1cm}{\centering 30.58 \\ \tiny $\pm$4.12}
        & \parbox[c]{1cm}{\centering 28.32 \\ \tiny $\pm$3.27}
        & \parbox[c]{1cm}{\centering 15.42 \\ \tiny $\pm$2.35}
        & \parbox[c]{1cm}{\centering 44.27 \\ \tiny $\pm$3.53} \\

        \midrule[0.1pt]
        
        \rowcolor{cyan!15}
        POWER (Ours) 
        & \parbox[c]{1cm}{\centering \textbf{65.74} \\ \tiny \textbf{$\pm$5.11}}
        & \parbox[c]{1cm}{\centering \textbf{28.10} \\ \tiny \textbf{$\pm$4.95}}
        & \parbox[c]{1cm}{\centering \textbf{54.47} \\ \tiny \textbf{$\pm$5.21}}
        & \parbox[c]{1cm}{\centering \textbf{28.94} \\ \tiny \textbf{$\pm$3.19}}
        & \parbox[c]{1cm}{\centering \textbf{36.19} \\ \tiny \textbf{$\pm$3.17}}
        & \parbox[c]{1cm}{\centering \textbf{30.52} \\ \tiny \textbf{$\pm$4.22}}
        & \parbox[c]{1cm}{\centering \textbf{43.26} \\ \tiny \textbf{$\pm$2.94}}
        & \parbox[c]{1cm}{\centering \textbf{12.59} \\ \tiny \textbf{$\pm$3.03}}
        & \parbox[c]{1cm}{\centering \textbf{89.17} \\ \tiny \textbf{$\pm$2.54}}
        & \parbox[c]{1cm}{\centering \textbf{13.03} \\ \tiny \textbf{$\pm$4.53}}
        & \parbox[c]{1cm}{\centering \textbf{40.27} \\ \tiny \textbf{$\pm$3.22}}
        & \parbox[c]{1cm}{\centering \textbf{35.24} \\ \tiny \textbf{$\pm$2.98}}
        & \parbox[c]{1cm}{\centering \textbf{35.79} \\ \tiny \textbf{$\pm$4.15}}
        & \parbox[c]{1cm}{\centering \textbf{18.13} \\ \tiny \textbf{$\pm$3.24}}
        & \parbox[c]{1cm}{\centering \textbf{26.54} \\ \tiny \textbf{$\pm$2.35}}
        & \parbox[c]{1cm}{\centering \textbf{16.23} \\ \tiny \textbf{$\pm$2.26}} \\
        \bottomrule[1pt]
    \end{tabular}
    }}
\end{table*}

\subsection{Performance Comparison (\textit{Answer for} \textbf{Q1})}
\label{sec: effectiveness}

To answer \textbf{Q1}, we present a performance comparison of our proposed POWER framework against various baselines in Table.~\ref{tab: compare baseline}. As shown, POWER consistently outperforms all baselines across both AM and FM metrics. We provide a detailed discussion of the reasons behind the sub-optimal performance of each baseline category.

\vspace{0.13cm}
\noindent \textbf{Comparison to FL/FGL Fine-tuning.} These methods lack targeted designs to alleviate LGF across tasks, resulting in poor retention of prior performance and consistently low AM and high FM. Algorithms originally designed for FGL (e.g., FedSage+, Fed-PUB, FedGTA) often perform worse than the simpler FedAvg, due to their complex focus on current tasks, which can lead to overfitting and worsen forgetting of previous tasks as local graphs evolve.

\vspace{0.13cm}
\noindent \textbf{Comparison to Federated CGL.} These methods combine the centralized CGL algorithm with FedAvg, mitigating LGF but not effectively addressing GEC. While they retain some memory of previous tasks, they still show a significant performance gap compared to POWER (e.g., Fed-ERGNN vs. POWER on Cora), mainly due to GEC, as discussed in Sec.~\ref{sec: empirical investigation}. Notably, FedTWP often matches FedAvg, losing most ability to retain previous task knowledge. We attribute this to its regularization of task-sensitive parameters, disrupted by the replacement of local parameters with the global model in FCGL.

\vspace{0.13cm}
\noindent \textbf{Comparison to Federated Continual Learning for CV.} These methods are tailored for federated continuous learning in computer vision but show significant shortcomings in the FCGL scenario. While they outperform most fine-tuning methods, they lag behind Fed-ERGNN and POWER. This is due to their heavy reliance on image data augmentation or the need for complex generative models to create pseudo images. In FCGL, simply removing data augmentation or using a basic GNN-based model doesn’t yield satisfactory results. Moreover, these methods fail to effectively address GEC.

\begin{table}
    \setlength{\abovecaptionskip}{0.13cm}
        \renewcommand{\arraystretch}{1.2}
      \centering
      \captionsetup{font={small,stretch=1}}
      \caption{\textbf{Ablation study results} on the Cora dataset.}
      \label{tab: ablation study}
      \resizebox{\linewidth}{20mm}{
    \setlength{\tabcolsep}{7mm}{
    \begin{tabular}{c|c|c}
    \toprule[1.0pt]
    Components & AM $\uparrow$ & FM $\downarrow$    \\
    \midrule[0.1pt]
    
    \rowcolor{cyan!15}
    POWER  & \parbox[c]{1cm}{\centering \textbf{65.74} \\ \tiny \textbf{$\pm$5.11}} & \parbox[c]{1cm}{\centering \textbf{28.10} \\ \tiny \textbf{$\pm$4.95}}\\
    
    \midrule[0.1pt]
    
    w/o LGF   & \parbox[c]{1cm}{\centering 43.42 \\ \tiny $\pm$2.27} & \parbox[c]{1cm}{\centering 74.25 \\ \tiny $\pm$3.34}\\
    w/o GEC    & \parbox[c]{1cm}{\centering 61.50 \\ \tiny $\pm$2.71} & \parbox[c]{1cm}{\centering 42.08 \\ \tiny $\pm$4.42}\\
    w/o LGF \& GEC     & \parbox[c]{1cm}{\centering 40.14 \\ \tiny $\pm$1.01} & \parbox[c]{1cm}{\centering 78.92 \\ \tiny $\pm$1.84}\\
    
    \midrule[0.1pt]
    
    \textit{Local CM}   & \parbox[c]{1cm}{\centering 63.22 \\ \tiny $\pm$4.27} & \parbox[c]{1cm}{\centering 30.21 \\ \tiny $\pm$3.14}\\
    \textit{Non-cumulative}  & \parbox[c]{1cm}{\centering 62.25 \\ \tiny $\pm$2.77} & \parbox[c]{1cm}{\centering 31.36 \\ \tiny $\pm$4.81}\\

    \bottomrule[1.0pt]
    \end{tabular}
    }}
    \end{table}

\subsection{Ablation Study (Answer for \textbf{Q2})} 
\label{sec: ablation study}

To address \textbf{Q2}, we analyze the composition of POWER, focusing on two key modules: \textbf{(1) LGF module}, which combines local-global coverage maximization (Eqs.~(\ref{eq: local-global embedding}-\ref{eq: coverage max})) with experience node replay (Eq.~(\ref{eq: local loss})); and \textbf{(2) GEC module}, which integrates pseudo prototype reconstruction (Eqs.~(\ref{eq: prototype encode}-\ref{eq: cumulative_label_distribution})) and trajectory-aware knowledge transfer (Eqs.~(\ref{eq: knn}-\ref{eq: div loss})). To evaluate each component's contribution, we conduct an ablation study on the Cora dataset. Additionally, we examine variations within each module: for the \textbf{LGF module}, we use local coverage maximization (\underline{\textit{Local CM}}), and for the \textbf{GEC module}, we replace the cumulative label distribution with a non-cumulative version (\underline{\textit{Non-cumulative}}). The results are presented in Table.~\ref{tab: ablation study}.

\vspace{0.15cm}
\noindent \textbf{Inter-module Ablation.} Removing the LGF Module causes significant performance drops of 22\% and 46\% in AM and FM, highlighting its critical role in FCGL's effectiveness. Excluding the GEC Module leads to smaller reductions of 4\% and 14\%, indicating it limits further performance gains but is less impactful than LGF. Removing both modules reduces POWER to a basic FedAvg algorithm.

\vspace{0.15cm}
\noindent \textbf{Intra-module Ablation.} Delving deeper, we examine the internal design choices of each module. The \textbf{Local CM} variant, which forgoes global insight during coverage maximization, leads to performance degradation, confirming that our local-global coverage maximization strategy benefits from global insights to select more representative experience nodes. Similarly, the \textbf{Non-cumulative} configuration weakens POWER's ability to encode graph evolution trajectory, confirming that the cumulative label distribution is key to preserving the evolving knowledge across local GNNs over tasks.

\vspace{0.15cm}
\textbf{In summary}, both these modules and their designed components play a pivotal role in shaping the overall performance.

\subsection{Sensitivity Analysis (Answer for \textbf{Q3})}
\label{sec: sensitivity}

\begin{figure*}[htb]
    \setlength{\abovecaptionskip}{0.2cm}
  \includegraphics[width=0.998\textwidth]{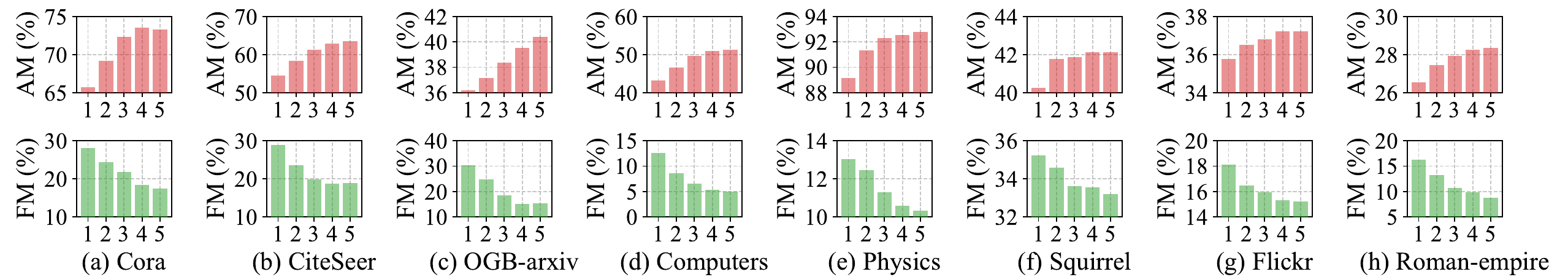}
  \captionsetup{font={small,stretch=1}}
  \caption{\textbf{Sensitivity analysis} for the replay buffer size $b$ across eight benchmark graph datasets.} 
\label{fig: sen_b}

\end{figure*}

\begin{figure}[htb]
\setlength{\abovecaptionskip}{0.2cm}
  \includegraphics[width=0.498\textwidth]{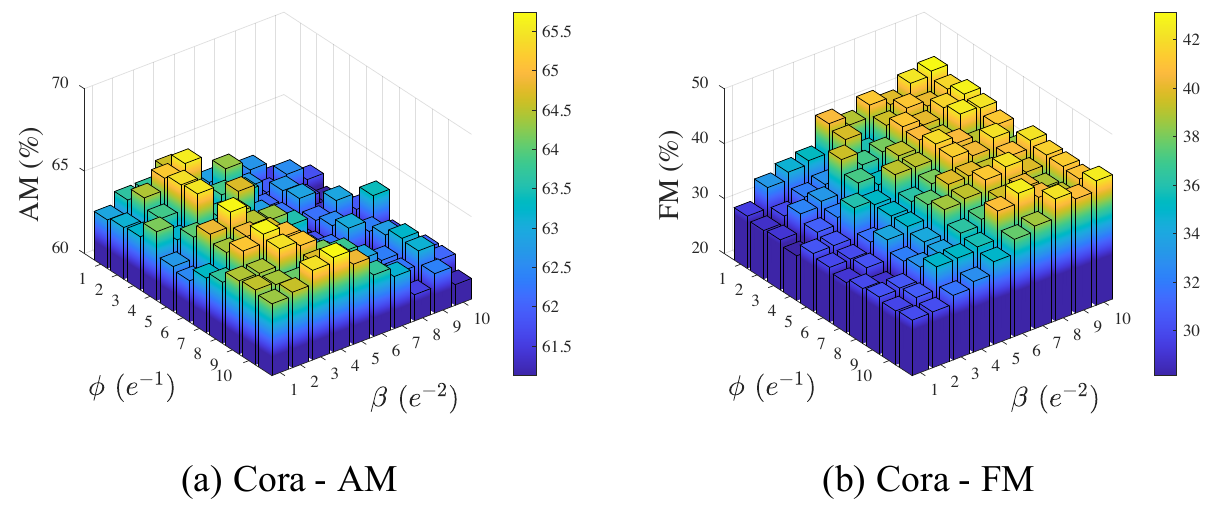}
  \captionsetup{font={small,stretch=1}}
  \caption{\textbf{Sensitivity analysis} for various combinations of the trade-off parameters $\beta$ and the decay coefficient $\phi$ on the Cora dataset.}
\label{fig: sen_comb}

\end{figure}

To address \textbf{Q3}, we perform a sensitivity analysis of hyperparameters in the POWER framework. \underline{First}, we examine the impact of replay buffer size \( b \) (Eq.~(\ref{eq: coverage max})), as shown in Fig.~\ref{fig: sen_b}. As observed, larger \( b \) improves both AM and FM. Our default \( b=1 \) minimizes storage costs, but in scenarios with sufficient storage, increasing \( b \) can enhance performance. \underline{Next}, we analyze the sensitivity of POWER to the trade-off parameter \( \beta \) (Eq.~(\ref{eq: local loss})) and decay coefficient \( \phi \) (Eq.~(\ref{eq: cumulative_label_distribution})), both affecting task knowledge retention. Fig.~\ref{fig: sen_comb} shows results on the Cora dataset, focusing on values near the optimal range. A large \( \beta \) may underplay old task knowledge, while a small \( \phi \) may fail to capture it effectively. POWER performs stably with appropriate \( \beta \) and \( \phi \) combinations, maintaining holistic performance.

\section{Related Work}
    \vspace{0.15cm}
    \textbf{Continual Graph Learning (CGL).} CGL introduces a paradigm where GNNs learn from evolving graphs while retaining prior knowledge. The main challenge, catastrophic forgetting, causes performance degradation on past tasks as GNNs adapt to new data. CGL research is mainly categorized into three approaches: (1) \textbf{Experience Replay} selects and stores a minimal set of data from previous tasks, aiming to efficiently retain representative data within memory constraints. This involves deciding which data to replay and how to use it for future tasks. Early works like ERGNN~\cite{zhou2021ergnn} store experience data at the node level, using strategies such as coverage and social influence maximization, and incorporating node classification losses during new task training. DSLR~\cite{choi2024dslr} refines node selection with a diversity-based strategy and improves node topology via graph structure learning. Recent studies~\cite{zhang2022cgl_ssm, zhang2023cgl_ssm_curvature, zhang2024pdgnns, han2024cgl_taco, liu2023cgl_cat, li2024cgl_gsip} focus on experience data selection at the subgraph or ego-network level, preserving topological information and achieving good performance. (2) \textbf{Parameter Regularization} alleviates catastrophic forgetting by adding a regularization term during new task training to preserve parameters crucial for prior tasks. For example, TWP~\cite{liu2021twp} introduces a regularization term to preserve key GNN parameters for node semantics and topology aggregation. To address forgetting caused by structural shifts, SSRM~\cite{su2023cgl_ssrm_regularization} introduces a regularization-based mitigation strategy. (3) \textbf{Architecture Isolation} allocates dedicated GNN parameters to each task, preventing interference when learning new tasks. Studies~\cite{rakaraddi2022cgl_architecture_iso_1, yoon2017cgl_architecture_iso_2} expand the GNN architecture to accommodate new graphs when its capacity is insufficient.Moreover, TPP~\cite{niu2024cgl_tpp} leverages Laplacian smoothing to profile graph tasks and learns a separate prompt-based classifier for each, thereby eliminating the need for replay and mitigating catastrophic forgetting in CGL.

    \vspace{0.15cm}
    \noindent \textbf{Federated Graph Learning (FGL).} Motivated by the success of federated learning in computer vision and natural language processing~\cite{yang2019fl_survey} and the demand for distributed graph learning, FGL has gained increasing attention. From the data and task perspectives, FGL studies are categorized into two settings: (1) \textbf{Graph-FL}, where each client collects multiple graphs for graph-level tasks, like graph classification. The main challenge is avoiding interference between clients' graph datasets, especially in multi-domain settings. For example, GCFL+~\cite{xie2021gcfl} introduces a GNN gradient pattern-aware technique for dynamic client clustering to reduce conflicts from structural and feature heterogeneity. (2) \textbf{Subgraph-FL}, where each client holds a subgraph of a global graph for node-level tasks like node classification. The key challenges are \textit{subgraph heterogeneity} and \textit{missing edges}. Fed-PUB~\cite{baek2022fedpub} addresses heterogeneity by enhancing local GNNs with random graph embeddings and personalized sparse masks for selective aggregation. FedGTA~\cite{li2024fedgta} encodes topology into smoothing confidence and graph moments to improve model aggregation. Other studies~\cite{li2024adafgl, huangfgl_fgssl, wan2024fgl_fggp, zhu2024fedtad} also achieve strong results on this challenge. To address missing edges, FedSage+~\cite{zhang2021fedsage} integrates node representations, topology, and labels across subgraphs, training a neighbor generator to restore missing links and achieve robust subgraph-FL. Other works~\cite{chen2021fedgl, yao2024fgl_fedgcn, zhang2024fgl_feddep} also excel in this area. Detailed insights into FGL research are available in surveys~\cite{fu2022fgl_survey_1, zhang2021fgl_survey_2, fu2023privacy_gml_survey} and benchmark studies~\cite{he2021fedgraphnn, WangFedScope_22_fsg, li2024openfgl}.

\vspace{0.1cm}
\section{Conclusion}
This paper pioneers the exploration of federated continual graph learning (FCGL) for node classification, bridging the gap between idealized centralized continual graph learning (CGL) setups and real-world decentralized challenges.  Through empirical analysis, we investigate the data characteristics, feasibility, and effectiveness of FCGL, identifying two critical challenges: local graph forgetting (LGF) and global expertise conflict (GEC). To address these, we propose a novel FCGL framework named POWER to mitigate LGF by selectively replaying experience nodes with maximum local-global coverage, and resolve GEC through pseudo-prototype reconstruction and trajectory-aware knowledge transfer.

\section*{Acknowledgments}
This work was supported by the Shenzhen Science and Technology Program (Grant No. KJZD20230923113901004), and the Fundamental Research Funds for the Central Universities, Sun Yat-sen University under Grant 24lgqb020-2.

\newpage
\bibliographystyle{ACM-Reference-Format}
\balance
\bibliography{FCGL_citation}

\appendix

\newpage
\clearpage
\section{Experiment Details}
\label{appendix: exp setting}

In this section, we detail our experiment setup. First, we present 10 baselines in \ref{appendix: baseline descriptions}. Then, we provide information on the model architecture, hyperparameters, and experiment environment in \ref{appendix: model architecture}, \ref{appendix: hyperparameters}, and \ref{appendix: experiment environment}, respectively.

\subsection{Dataset Descriptions} 
\label{appendix: dataset description}
We evaluate 8 graph datasets across 6 domains:

\textbf{Cora} and \textbf{CiteSeer}~\cite{Yang16cora} are citation networks with nodes as papers and edges as citations. Features are binary word vectors. Both are widely used for node classification.

\textbf{OGB-Arxiv}~\cite{hu2020ogb,wang2020microsoft_mag} is a large citation graph from MAG, where each node is a paper with features from averaged skip-gram embeddings of its title and abstract.

\textbf{Computers}~\cite{shchur2018amazon_datasets} is an Amazon co-purchase graph. Nodes are products, edges indicate frequent co-purchases, and features are bag-of-words from reviews.

\textbf{Physics}~\cite{shchur2018amazon_datasets,wang2020microsoft_mag} is a MAG co-authorship graph, where nodes are authors and edges denote collaborations. Features are extracted from publication keywords; labels represent research fields.

\textbf{Squirrel}~\cite{pei2020geomgcn} is a Wikipedia page graph, with nodes as pages linked via mutual hyperlinks. Features are key nouns, and labels are based on traffic levels.

\textbf{Roman-Empire}~\cite{platonov2023hete_gnn_survey4,lhoest2021english_wiki} is a word-level graph from a Wikipedia article. Nodes are words, and edges follow word order or syntactic dependencies. Used for node classification.

\subsection{Baseline Descriptions} 
\label{appendix: baseline descriptions}

Since this paper pioneers the study of FCGL, there are no established baselines specific to this setting. To ensure a meaningful comparison, we adapt representative methods from closely related domains. In total, we evaluate 10 baseline methods, which are systematically grouped into three distinct categories based on their methodological principles. The detailed descriptions of these baseline methods are provided as follows:

\vspace{0.16cm}
\noindent \textbf{(1) FL/FGL Fine-tuning.} This category includes FedAvg~\cite{mcmahan2017fedavg}, FedSage+~\cite{zhang2021fedsage}, Fed-PUB~\cite{baek2022fedpub} and FedGTA~\cite{li2024fedgta}, which simply replace the local client training process of the FL/FGL algorithm with fine-tuning on sequential tasks to adapt to FCGL settings.

 \textbf{FedAvg}~\cite{mcmahan2017fedavg} is a simple yet effective method in FL enabling decentralized model training while preserving data privacy. A central server distributes the global model to clients for local updates. The server then aggregates the clients' models to form a new global model, which is used to update local models in the next round.

 \textbf{FedSage+}~\cite{zhang2021fedsage} integrates node features, link structures, and labels using a GraphSAGE~\cite{hamilton2017graphsage} model with FedAvg~\cite{mcmahan2017fedavg} for FGL over local subgraphs. It also introduces a neighbor generator to handle cross-client missing links, improving robustness and ensuring a more comprehensive graph representation. This enhances the model's generalization across clients in FGL.

 \textbf{Fed-PUB}~\cite{baek2022fedpub} is a framework for personalized subgraph-FL that improves local GNNs independently instead of using a global model. It computes similarities between local GNNs with functional embeddings from random graph inputs, enabling weighted averaging for server-side aggregation. A personalized sparse mask at each client selectively updates subgraph-relevant parameters.

 \textbf{FedGTA}~\cite{li2024fedgta} combines large-scale graph learning with FGL. Clients encode topology and node attributes, compute local smoothing confidence and mixed moments of neighbor features, and upload them to the server. The server then aggregates models personalized using local smoothing confidence as weights.

\vspace{0.16cm}
\noindent \textbf{(2) Federated CGL.} This category includes Fed-ERGNN, FedTWP, and Fed-DSLR, each combining the FedAvg~\cite{mcmahan2017fedavg} algorithm with a representative CGL method, ERGNN~\cite{zhou2021ergnn}, TWP~\cite{liu2021twp}, and DSLR~\cite{choi2024dslr}, respectively. In each case, local training leverages the corresponding CGL technique to handle evolving local graphs, while client-server communication follows the standard FedAvg process.

 \textbf{Fed-ERGNN} is the federated variant of ERGNN~\cite{zhou2021ergnn}. ERGNN is a representative CGL approach based on the experience replay mechanism, which stores experience data at the node level, using strategies such as coverage and social influence maximization, and incorporating node classification losses during new task training.

 \textbf{Fed-TWP} is the federated variant of TWP~\cite{liu2021twp}. As an effective parameter regularization-based CGL technique, TWP preserves key GNN parameters for node semantics and topology aggregation by constraining updates through regularization.

 \textbf{Fed-DSLR} is the federated variant of DSLR~\cite{choi2024dslr}. DSLR is another experience replay-based CGL technique, which refines node selection with a diversity-based strategy and improves node topology via graph structure learning. 

\vspace{0.16cm}
\noindent \textbf{(3) Federated Continual Learning for CV.} This category includes GLFC~\cite{dong2022glfc}, TARGET~\cite{zhang2023target}, and LANDER~\cite{tran2024lander}, which are designed for federated continual learning in computer vision. Due to format differences between images and graphs, some modules and workflows need adaptation for FCGL.

 \textbf{GLFC}~\cite{dong2022glfc} is the first method to learn a global class-incremental model in FL, addressing both local and global catastrophic forgetting. It introduces a prototype communication mechanism to generate images on the server side, using them as a proxy evaluation dataset to select the historical global model with the best performance on previous tasks.

 \textbf{TARGET}~\cite{zhang2023target} alleviates catastrophic forgetting by transferring knowledge of old tasks to new ones via the global model. A generator also creates synthetic images to simulate the global data distribution, eliminating the need to store real data.

\textbf{LANDER}~\cite{tran2024lander} is a generative model-based method for federated continual learning. It uses label text embeddings (LTE) from pre-trained language models as anchor points in server-side, data-free knowledge transfer, constraining feature embeddings to mitigate catastrophic forgetting.

\subsection{Model Architecture}
\label{appendix: model architecture}
 We employ a 2-layer GAT~\cite{velivckovic2017gat} with 64 hidden units as the backbone of clients and central server. Notably, model-specific baselines like FedSage+~\cite{zhang2021fedsage} adhere to the custom architectures specified in their original papers. For the POWER framework, the gradient encoding network and prototype reconstruction network are implemented as 4-layer MLPs with 128, 128, and 64 hidden units. To ensure compatibility with graph data or matrix inputs, we replace the visual models used in the baseline comparisons with their respective equivalent layers of GAT or MLP.

\subsection{Hyperparameters}
\label{appendix: hyperparameters}

\begin{figure*}[h]
  \includegraphics[width=0.998\textwidth]{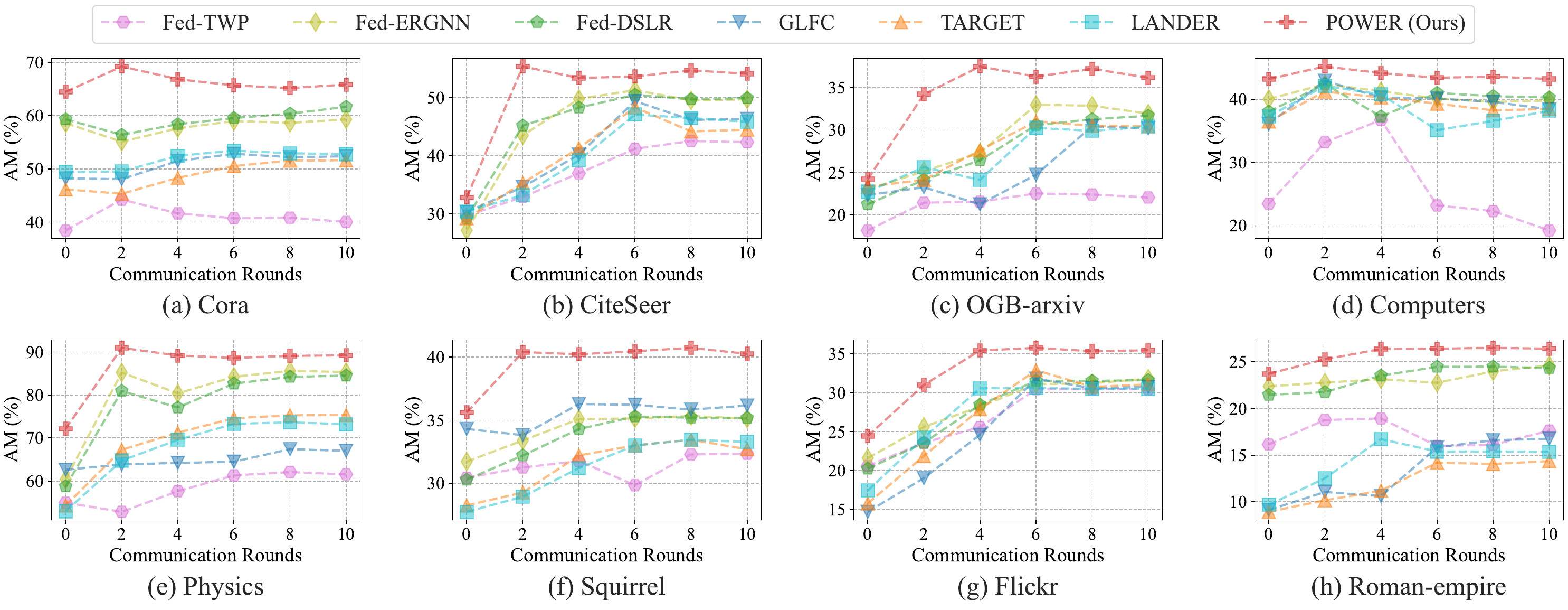}
  \captionsetup{font={small,stretch=1}}
  \caption{\textbf{Convergence curves} for the proposed POWER framework and six baseline methods across eight benchmark graph datasets \textbf{observed during training on the final task}, where ``Round 0'' corresponds to the state prior to initiating training on the final task.}
\label{fig: convergence}
\end{figure*}

In each task, we conduct 10 communication rounds, each comprising 3 local training epochs. These epochs utilize the Adam optimizer \cite{kingma2014adam_optimizer} with hyperparameters set to a learning rate of $1 \times 10^{-2}$, weight decay of $5 \times 10^{-4}$, and a dropout rate of 0.5. For all considered baselines, we adopt hyperparameters from the original publications whenever possible. In cases where these are not specified, we employ an automated hyperparameter optimization using the Optuna framework \cite{akiba2019optuna}. For the POWER method, specific hyperparameters are fixed as follows: trade-off parameter $\alpha$ at 0.5, number of experience nodes per task $b$ at 1, and the K-Nearest Neighbor parameter at 1, as per Eqs.~(\ref{eq: local-global embedding}), (\ref{eq: coverage max}), and (\ref{eq: knn}), respectively. Exploratory hyperparameters include the the threshold distance $\eta$ set within $\{0.01, 0.1, 0.5\}$ in Eq.~(\ref{eq: local-global coverage}), trade-off parameter $\beta$ in Eq.~(\ref{eq: local loss}) and the past task decay coefficient $\phi$ varied from 0 to 1 in increments of $0.01$ in Eq.~(\ref{eq: cumulative_label_distribution}). Finally, the pseudo prototype reconstruction process in 
Eq.~(\ref{eq: gradient match loss}) is optimized with 300 iterations using the LBFGS optimizer \cite{liu1989lbfgs_optimizer}, which has an initial learning rate of 1.0. Results are reported as the mean and variance across 10 standardized runs.

\begin{table}
  \setlength{\abovecaptionskip}{0.3cm}
  \centering
  \captionsetup{font={small,stretch=1}}
  \caption{\textbf{Theoretical communication overhead} comparison of POWER and FedAvg. Notably, \textbf{FedAvg represents the lowest communication cost} of mainstream federated learning processes.}
  \label{tab: theoretical overhead}
  \resizebox{\linewidth}{12mm}{
\setlength{\tabcolsep}{1.5mm}{
\begin{tabular}{c|cc}
\toprule[1.0pt]
Method    & Content & Overhead    \\
\midrule[0.1pt]
FedAvg      & \makecell{GNN Parameters}    & $\mathcal{O}(KTRN_{\mathbf{\Theta}_\text{GNN}})$       \\
\midrule[0.1pt]
POWER (Ours)      & \makecell{GNN Parameters\\Prototype Gradients}    & \makecell{$\mathcal{O}(KTRN_{\mathbf{\Theta}_\text{GNN}} +$\\ $KCN_{\mathbf{\Theta}_\text{GE}})$}       \\
\bottomrule
\end{tabular}
}}
\vspace{-3mm}
\end{table}

\begin{table}
  \setlength{\abovecaptionskip}{0.3cm}
  \centering
  \captionsetup{font={small,stretch=1}}
  \caption{Running time comparison of POWER and baselines.}
  \label{tab: running time}
  \resizebox{\linewidth}{25mm}{
\setlength{\tabcolsep}{1.2mm}{
\begin{tabular}{lcc}
\toprule
Method & Client Execution (s) & Server Execution (s) \\
\midrule
FedAvg         & 0.0519 & 0.0005 \\
FedSage+       & 0.7589 & 0.0005 \\
Fed-PUB        & 0.0878 & 0.0029 \\
FedGTA         & 0.0717 & 0.0312 \\
Fed-TWP        & 0.0694 & 0.0005 \\
Fed-ERGNN      & 0.0569 & 0.0005 \\
Fed-DSLR       & 0.1104 & 0.0005 \\
GLFC           & 0.1786 & 3.9241 \\
TARGET         & 0.0783 & 3.1407 \\
LANDER         & 0.0812 & 1.5676 \\
POWER (ours)   & 0.0632 & 1.9388 \\
\bottomrule
\end{tabular}
}}
\end{table}

\subsection{Experiment Environment}
\label{appendix: experiment environment}

    The experimental machine is an Intel(R) Xeon(R) Gold 6240 CPU @ 2.60GHz and NVIDIA A100 with 80GB memory and CUDA 12.4.
    The operating system is Ubuntu 22.04.5 with 251GB memory.

\section{Additional Experiments}

In this section, we conduct additional experiments to answer these questions: \textbf{Q4}: Does POWER demonstrate superiority in terms of efficiency (\ref{appendix: efficiency})? and \textbf{Q5}: How does POWER perform under sparse settings like missing features/edges/labels and low client participation rate (\ref{appendix: sparsity robustness})?

\subsection{Efficient Analysis (Answer for \textbf{Q4})}
\label{appendix: efficiency}

To address \textbf{Q4}, we provide a detailed analysis of convergence speed, communication overhead, and runing time, as outlined below.

\begin{figure*}[h]
  \includegraphics[width=0.998\textwidth]{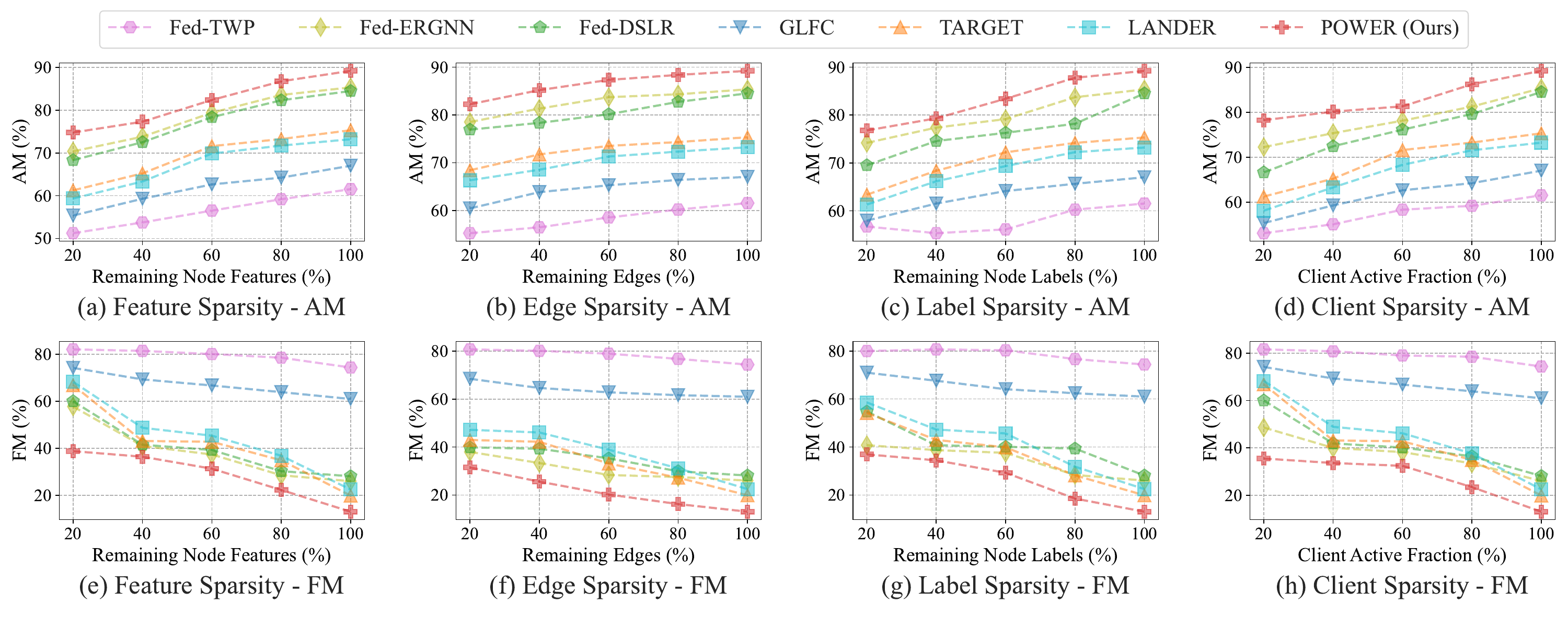}
  \captionsetup{font={small,stretch=1}}
  \caption{\textbf{Performance under sparse settings} for the proposed POWER framework and six baseline methods on the Physics dataset.}
\label{fig: sparse performance}
\end{figure*}

\vspace{0.1cm}
\noindent \textbf{Convergence Speed.} We present the convergence curves of POWER alongside six continual learning-related baselines: Fed-TWP, Fed-ERGNN, Fed-DSLR, GLFC, TARGET, and LANDER. Notably, methods such as \textbf{FL/FGL Fine-tuning} are excluded from this comparison, as their limitations in FCGL scenarios have already been established in Sec.~\ref{sec: effectiveness}. The results, illustrated in Fig.~\ref{fig: convergence}, demonstrate that POWER consistently converges faster than all baselines, showcasing its enhanced efficiency. This efficiency is critical for real-world applications, as it requires fewer communication rounds to achieve optimal performance, ultimately reducing communication overhead significantly. We attribute this to the effectiveness of the pseudo prototype reconstruction module in capturing localized expertise from each client. Additionally, the trajectory-aware knowledge transfer mechanism enables the global model to assimilate this expertise more effectively. These factors contribute to POWER's ability to achieve fast convergence.

\vspace{0.1cm}
\noindent \textbf{Communication Overhead.} To assess POWER's practicality in real-world settings, we compare its theoretical communication cost with the baseline FedAvg, which also underpins Fed-ERGNN, Fed-TWP, and Fed-DSLR, representing the minimal communication cost among mainstream federated learning approaches. Table.~\ref{tab: theoretical overhead} summarizes the results, where $K, T, R, C$ represent the number of clients, tasks, communication rounds, and total classes across all tasks, respectively. Additionally, $N_{\mathbf{\Theta}_\text{GNN}}$ and $N_{\mathbf{\Theta}_\text{GE}}$ denote the parameter counts for the local GNN and gradient encoding network. Both POWER and FedAvg require transmitting GNN parameters to the server, incurring a communication cost of $\mathcal{O}(KTRN_{\mathbf{\Theta}_\text{GNN}})$. POWER introduces an additional term for the prototype gradient, sent from clients to the server, with a cost of $\mathcal{O}(KCN_{\mathbf{\Theta}_\text{GE}})$. Importantly, in practical settings, $\mathcal{O}(KCN_{\mathbf{\Theta}_\text{GE}})$ is typically \underline{\textit{far smaller}} than $\mathcal{O}(KTRN_{\mathbf{\Theta}_\text{GNN}})$. This reduction can be attributed to several natural factors: \textbf{(1)} Real-world graph data tends to be complex and noisy, often requiring a large number of communication rounds $R$ for convergence, while POWER only uploads prototype gradients in the first round of communication for each task. \textbf{(2)} Due to the personalized nature of local data sources and collection methods, each client typically lacks data for some categories and, as a result, does not transmit prototypes for all classes $C$. \textbf{(3)} The GNN models in practice have significantly larger parameter counts compared to our lightweight gradient encoding network (i.e., $N_{\mathbf{\Theta}_\text{GE}} \ll N_{\mathbf{\Theta}_\text{GNN}}$).

\vspace{0.1cm}
\noindent \textbf{Runing Time.} To further evaluate the efficiency of POWER, we measure its runtime against baseline methods on the Cora dataset. The results are summarized in Table.~\ref{tab: running time}. Notably, (1) the single-round client execution time of POWER is lower than that of most baselines; (2) the single-round server execution time of POWER is shorter than that of federated continual learning methods for vision tasks (i.e., GLFC, TARGET, LANDER). Although it exceeds the runtime of a simple federated variant of centralized CGL, this additional overhead enables POWER to effectively mitigate GEC.

\subsection{Sparsity Robustness (Answer for \textbf{Q5})}
\label{appendix: sparsity robustness}

In response to \textbf{Q5}, we conducted experiments to assess the resilience of the POWER framework relative to six baselines across various sparsity conditions: (1) \underline{\textit{feature sparsity}}, where node features are selectively obscured to mimic incomplete information; (2) \underline{\textit{edge sparsity}}, which simulates partial graph connectivity by randomly removing edges; (3) \underline{\textit{label sparsity}}, where only a subset of nodes have available labels to represent scenarios with incomplete supervision; and (4) \underline{\textit{client sparsity}}, characterized by limited client participation in federated communication due to factors such as network constraints or hardware limitations. These sparsity settings reflect the limitations encountered in real-world FCGL scenarios.

\vspace{0.1cm}
\noindent \textbf{Sparse Feature/Edge/Label.} Fig.~\ref{fig: sparse performance} presents the results for sparse feature/edge/label settings, where panels (a)-(c) present the AM metric, and (e)-(g) present the FM metric. As observed, POWER reveals superior performance across all degrees of sparsity. We attribute this superiority to its pseudo prototype reconstruction mechanism and efficient knowledge transfer processes. Specifically, for \underline{\textit{feature sparsity}}, POWER calculates prototypes using node features within the same category, effectively compensating for missing features by utilizing the complementary data from multiple nodes. For \underline{\textit{edge sparsity}}, although POWER does not specifically target missing connections, its superior performance likely stems from the strong overall learning architecture and effective knowledge transfer process. For \underline{\textit{label sparsity}}, POWER's server-side knowledge transfer process allows each client to share its local insights in a privacy-preserving manner, which enhances the global model's ability to compensate for sparse local supervision, effectively learning from limited label information.

\vspace{0.1cm}
\noindent \textbf{Sparse Client Participation}. The results for \underline{\textit{client sparsity}} settings are depicted in Fig.~\ref{fig: sparse performance}, where panel (d) presents the AM metric, and (h) presents the FM metric. As observed, POWER maintains stable performance across various participation ratios. We attribute this robustness in client-sparse settings to its effective prototype reconstruction mechanism, detailed in the methods section. Specifically, when POWER receives prototype gradients from a participating client, it effectively reconstructs and stores these prototypes in a global buffer. Thus, even if a client discontinues participation in subsequent federation rounds, the knowledge from its prototypes continuously contributes to the knowledge transfer process, thereby optimizing the global model.

\textbf{In summary}, in the FCGL context, POWER outperforms traditional CGL methods and vision-focused federated learning algorithms, demonstrating superiority for sparsity challenges.

\end{document}